\pdfoutput=1

\documentclass[11pt]{article}

\usepackage[]{acl}

\usepackage{times}
\usepackage{latexsym}
\usepackage{adjustbox}
\usepackage{booktabs}
\usepackage[nameinlink,capitalize]{cleveref}
\usepackage{dirtytalk}
\usepackage{multicol,multirow}
\usepackage{hyperref}
\usepackage{multirow}
\usepackage{tikz}
\usepackage{pgfplots}
\pgfplotsset{width=10cm,compat=1.17}
\usepackage{xcolor}
\usepackage{verbatim}

\usepackage[T2A,T1]{fontenc}

\usepackage{CJKutf8}
\newcommand{\zh}[1]{\begin{CJK}{UTF8}{bsmi}#1\end{CJK}}

\usepackage[utf8]{inputenc}

\usepackage{microtype}

\usepackage{inconsolata}

\usepackage{subcaption}
\usepackage{dirtytalk}  %

\newcommand{\nerexample}[2]{%
  \underline{#1}$_{\scriptsize\texttt{#2}}$%
}

\newcommand{\shortname}[0]{UNER}

\title{Universal NER: A Gold-Standard Multilingual \\Named Entity Recognition Benchmark}

\author{\bf Stephen Mayhew$^{\alpha}$ \quad Terra Blevins$^{\beta}$ \quad Shuheng Liu$^{\gamma}$ \quad Marek Šuppa$^{\delta}$$^{\epsilon}$ \\ 
\bf Hila Gonen$^{\beta}$ \quad Joseph Marvin Imperial$^{\zeta,\eta}$ \quad Börje F. Karlsson$^{\theta}$ \quad Peiqin Lin$^{\iota}$\\ 
\bf Nikola Ljubešić$^{\kappa}$ \quad LJ Miranda$^{\lambda}$ \quad Barbara Plank$^{\iota,\mu}$ \quad Arij Riabi$^{\nu,\xi}$ \quad Yuval Pinter$^{o}$ \\
$^{\alpha}$Duolingo, 
$^{\beta}$University of Washington, 
$^{\gamma}$Georgia Institute of Technology, \\
$^{\delta}$Comenius University in Bratislava,
$^{\epsilon}$Cisco,
$^{\zeta}$National University Philippines, \\
$^{\eta}$University of Bath,
$^{\theta}$Beijing Academy of Artificial Intelligence, \\
$^{\iota}$LMU Munich, $^{\kappa}$Jožef Stefan Institute,
$^{\lambda}$Allen Institute for Artificial Intelligence, \\
$^{\mu}$IT University of Copenhagen,
$^{\nu}$Inria Paris,
$^{\xi}$Sorbonne Université,
$^{o}$Ben-Gurion University \\
\url{stephen@duolingo.com} \quad \quad \url{blvns@cs.washington.edu}}

\begin{document}
\maketitle
\begin{abstract}

We introduce Universal NER (\shortname{}), an open, community-driven project to develop gold-standard NER benchmarks in many languages. The overarching goal of \shortname{} is to provide high-quality, cross-lingually consistent annotations to facilitate and standardize multilingual NER research. \shortname{} v1 contains 19 datasets annotated with named entities in a cross-lingual consistent schema across 13 diverse languages. In this paper, we detail the dataset creation and composition of \shortname{}; we also provide initial modeling baselines on both in-language and cross-lingual learning settings. We will release the data, code, and fitted models to the public.\footnote{\url{https://www.universalner.org}, UNER v1 available at \url{https://doi.org/10.7910/DVN/GQ8HDL}} 

\end{abstract} 

\section{Introduction}
\label{sec:intro}
High-quality data in many languages is necessary for broadly multilingual natural language processing. In named entity recognition (NER), the majority of annotation efforts are centered on English, and cross-lingual transfer performance remains brittle~\citep[e.g.,][]{chen2023better,ma2023colada}. %
Amongst non-English human-annotated NER datasets, while there have been multiple separate efforts in this front ~\cite[e.g.,][]{agic-ljubesic-2014-setimes,plank-2019-neural,adelani-etal-2022-masakhaner}, these either have disjoint annotation schemes and labels, cover a single language or small set of related languages, or are not widely accessible~\cite[e.g.,][]{strassel-tracey-2016-lorelei}. 
For most of the world's languages, the only readily available NER data is the automatically annotated WikiANN dataset~\cite{pan-etal-2017-cross}, though this annotation paradigm introduces data quality issues and limits its usefulness for evaluation~\cite{lignos-etal-2022-toward}.

\begin{figure}
    \centering
    \includegraphics[width=\linewidth]{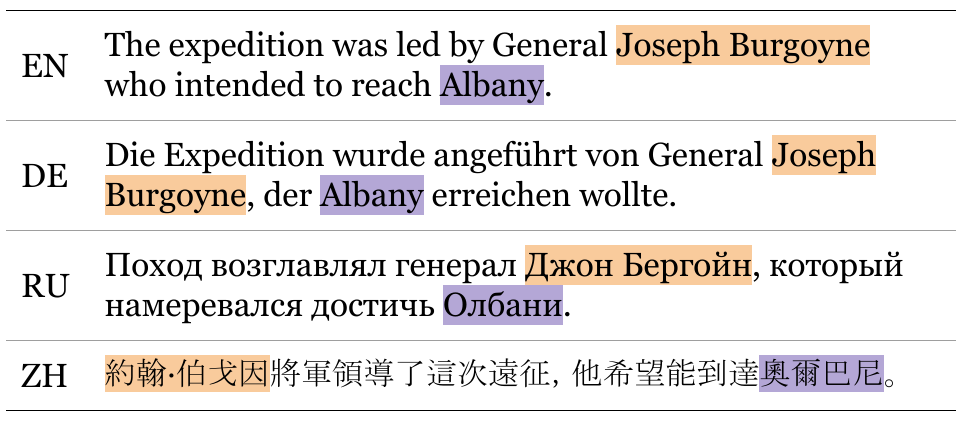}
    \caption{Parallel sentences annotated with \textcolor[HTML]{e69138}{person (\texttt{PER})} and \textcolor[HTML]{8e7cc3}{location (\texttt{LOC})} named entities in English (EN), German (DE), Russian (RU), and Chinese (ZH).}
    \label{fig:examples}
\end{figure}

To address this data gap, we propose Universal NER (\shortname), an open community effort to develop gold-standard named entity recognition benchmarks across many languages. Each dataset in Universal NER is annotated by primarily native speakers on the text of an existing Universal Dependencies treebank~\cite[UD;][]{nivre-etal-2020-universal}.
Inspired by Universal Dependencies, the overarching philosophy of the \shortname{} project is to provide a shared, universal definition, tagset, and annotation schema for NER that is broadly applicable across languages (\autoref{fig:examples}).

The current version of Universal NER, \shortname{}~v1, contains 19 datasets spanning 13 languages (Section \ref{sec:stats}).
To establish performance baselines on \shortname{}, we finetune an XLM-R model on various training configurations (Section~\ref{sec:baselines})and show that while NER transfer performance between European languages is relatively strong, there remains a gap when transferring to different scripts or language typologies.

The goal of the \shortname{} project is to facilitate multilingual research on entity recognition by addressing the need in the multilingual NLP community for standardized, cross-lingual, and manually annotated NER data. With the release of \shortname{} v1, we plan to expand \shortname{} to new languages and datasets, and we welcome all new annotators interested in developing the project.

\section{Dataset Design Principles}
\label{sec:tagset}
Named entity recognition (NER) is the task of identifying text spans in a given context that uniquely refer to specific \textit{named entities}. 
The task of NER has a long tradition~\cite{grishman_2019} and facilitates many downstream NLP applications, such as information retrieval \cite{khalid2008impact} and question answering \cite{molla-etal-2006-named}.
Furthermore, successful NER tagging requires a model to reason about semantic and pragmatic world knowledge, which makes the task an informative evaluation setting for testing NLP model capabilities.

As with Universal Dependencies, the goal of Universal NER is to develop an annotation schema that can work in any language. Traditionally, the UD \cite{nivre-etal-2016-universal} and UPOS \cite{petrov-etal-2012-universal} projects have chosen what amounts to the intersection of tags across all language-specific tagsets, keeping the resultant tagset broad and simple. We follow a similar strategy, picking tags that broadly cover the space of proper nouns.

Universal NER's annotation schema emphasizes three coarse-grained entity types: Person (\texttt{PER}), Organization (\texttt{ORG}), and Location (\texttt{LOC}).
We provide a short description and an example for each tag.

\paragraph{\texttt{PER}} The \texttt{PERSON} tag includes names of people, real or fictional, but not nominals.

``Mr. \nerexample{Robinson}{PER} smiled at the teacher.''

\paragraph{\texttt{ORG}} The \texttt{ORGANIZATION} tag is used for named collections of people.

``The \nerexample{FDA}{ORG} announced time travel pills tomorrow.''

\paragraph{\texttt{LOC}} The \texttt{LOCATION} tag covers all types of named locations. 

``I will arise and go now, and go to \nerexample{Innisfree}{LOC}''\\

\noindent \autoref{fig:examples} demonstrates how named entities and their corresponding annotations surface across languages. In some cases (such as in the English and German sentences), the surface forms of named entities are shared.
However, often these forms vary---as in the Russian and Chinese examples---which makes entity identification and tagging more challenging, particularly in cross-lingual settings.

\paragraph{Annotation Guidelines}
In preparation for annotation, we developed extensive annotation guidelines,\footnote{\url{http://www.universalner.org/guidelines/}} 
using the NorNE project guidelines \cite{jorgensen-etal-2020-norne} as a starting point. Along with tag descriptions, our guidelines include many examples, as well as instructions for dealing with ambiguity and unclear constructions, such as email addresses, pet names, and typographical errors.

We expect that the guidelines will be further refined and updated as annotation proceeds. To manage this, we track version numbers and changelogs for different iterations of the guidelines. Each data release will include the corresponding annotation guidelines at the time of release.

\section{Dataset Annotation Process}
\label{sec:annotation}
Having described the theoretical basis for the tagset, we now discuss the mechanics of annotation.

\paragraph{Sourcing Data}
We chose the Universal Dependency corpora as the default base texts for annotation. This jumpstarts the process: there is high coverage of languages, and the data is already collected, cleaned, tokenized, and permissively licensed. Further, by adding an additional annotation layer onto an already rich set of annotations, we not only support verification in our project (Section \ref{ssection:propn}) but also enable multilingual research on the full pipeline of core NLP. Since UD is annotated at the word level, we follow a BIO annotation schema (specifically IOB2), where words forming the beginning (inside) part of an \texttt{X} entity ($\texttt{X}\in \{$\texttt{PER}, \texttt{LOC}, \texttt{ORG}$\}$) are annotated \texttt{B-X} (\texttt{I-X}, respectively), and all other words are given an \texttt{O} tag. For the sake of continuity, we preserve all tokenization from UD.

While UD is the default data source for UNER, we do not limit the project to UD corpora (particularly for languages not currently included in UD). The only criterion for inclusion in the \shortname{} corpus is that the tagging schema matches the \shortname{} guidelines. We are also open to converting existing NER efforts on UD treebanks to \shortname{}. In this initial release, we include four datasets that are transferred from other manual annotation efforts on UD sources (for \textsc{da}, \textsc{hr}, \textsc{arabizi}, and \textsc{sr}).

\paragraph{Sourcing Annotators}
For the initial \shortname{} annotation effort, we recruited annotators from the multilingual NLP community through academic networks on social media. Annotators were organized via channels in a Slack workspace.
Annotators of the datasets included in \shortname{} thus far are unpaid volunteers. We expect that annotators are native speakers of their annotation language, or are highly proficient, but we did not issue any language tests. For the first release of \shortname{}, the choice of the 13 dataset languages is solely dependent on the availability of annotators. As the project continues, we expect that additional languages and datasets will be added as annotators in more languages become available to contribute.

\paragraph{Annotation Tool}
We collect annotations for the UD treebanks using TALEN~\citep{mayhew-roth-2018-talen}, a web-based tool for span-level sequence labeling.\footnote{\url{https://github.com/mayhewsw/talen-react}}
TALEN includes an optional feature that propagates annotations -- if the user annotates ``McLovin'' in one section of the document, every other instance of ``McLovin'' in that document is annotated as well. This significantly speeds up annotation but risks over-annotation mistakes. For example, consider the token ``US'', which may appear with different senses in contexts such as ``The US economy...'' or ``THEY OFFERED TO BUY US LUNCH!''

\paragraph{Secondary Annotators} 
In addition to collecting a complete set of annotations from a primary annotator for each dataset, we also gather secondary annotations from another annotator on (at least) a subset of the data in order to estimate inter-annotator agreement (Section~\ref{sec:iaa}).
We aim for at least 5\% coverage of each data split with these secondary annotations, although most datasets have significantly more (Table \ref{tab:iaa-scores}).
When a document has multiple annotators, we include the labels from the annotator with the most entities annotated in that document in the final dataset. This means a dataset may have multiple annotators, but each document has exactly one. We retain annotator identities in the data files.

\paragraph{Annotation Differences and Resolution}
When annotators disagreed on annotation decisions or the inter-annotator agreement scores were low, we encouraged them to discuss the disagreements and decide if they were conflicting interpretations of the guidelines or fundamental disagreements. In the former case, annotators came to an agreement on guideline interpretations and updated annotations accordingly. In the latter, the annotations were kept as-is. Not every dataset had this resolution process.

The multilingual nature of this process also highlighted cross-language differences in named entities that affect NER annotation. For instance, most languages in UNER use capitalization as a marker of proper nouns and, therefore, named entities. However, Chinese does not include capitalization in its script, which makes identifying named entities more difficult and time-consuming than in other languages, potentially leading to more annotation errors. Differences in annotating NER across languages also stem from divergent definitions of proper nouns (\texttt{PROPN}) by language and the effects of translation artifacts; these issues are discussed further in Sections \ref{ssection:propn} and \ref{ssection:pud-agreement}, respectively.

\paragraph{OTHER Tag}
As a helpful check for annotators, we allow the option of annotating a fourth entity type, Other (\texttt{OTH}), which is not included in the final dataset. This had several purposes: to store annotations that behaved like mentions, but didn't conform to the guidelines of the other tags; to measure potential annotation disagreement on ambiguous cases; and to store an additional layer of annotation.
Not all annotators used it, and those that did were sometimes inconsistent. In practice, \texttt{OTH} was most often applied to languages, nationalities, and brands.
The \texttt{OTH} tag roughly corresponds to the \texttt{MISC} tag used in CoNLL 2003, which has been described as being ``ill-defined''~\citep{adelani-etal-2022-masakhaner}.

\setlength{\tabcolsep}{4pt}
\begin{table*}[ht]
    \centering
    \scalebox{0.75}{
    \begin{tabular}{@{} llrrrr rrrr rrrr @{}}
    \toprule
     &  & \multicolumn{4}{c}{Sentences} & \multicolumn{4}{c}{Entities} & \multicolumn{4}{c}{Tokens} \\
     \cmidrule(lr){3-6} \cmidrule(lr){7-10} \cmidrule(lr){11-14}
     Lang. & Dataset & Train & Dev & Test & All & Train & Dev & Test & All & Train & Dev & Test & All \\
    \midrule
    \textsc{da} & \texttt{ddt} & 4,383 & 564 & 565 & 5,512 & 3,022 & 379 & 446 & 3,847 & 80,378 & 10,332 & 10,023 & 100,733 \\
    \textsc{en} & \texttt{ewt} & 12,543 & 2,001 & 2,077 & 16,621 & 7,022 & 966 & 1,088 & 9,076 & 204,579 & 25,149 & 25,097 & 254,825 \\
    \textsc{hr} & \texttt{set} & 6,914 & 960 & 1,136 & 9,010 & 8,261 & 1,218 & 1,403 & 10,882 & 152,857 & 22,292 & 24,260 & 199,409 \\
    \textsc{pt} & \texttt{bosque} & 7,018 & 1,172 & 1,167 & 9,357 & 8,101 & 1,401 & 1,215 & 10,717 & 171,776 & 28,447 & 27,604 & 227,827 \\
    \textsc{qaf} & \texttt{arabizi} & 1003 & 139 & 145 & 1287 & 1320 & 204 & 194 & 1718 & 15,522 & 2,124 & 2,118 & 19,764 \\
    \textsc{sk} & \texttt{snk} & 8,483 & 1,060 & 1,061 & 10,604 & 2,707 & 636 & 915 & 4,258 & 80,628 & 12,733 & 12,736 & 106,097 \\
    \textsc{sr} & \texttt{set} & 3,328 & 536 & 520 & 4,384 & 5,020 & 742 & 847 & 6,609 & 74,259 & 11,993 & 11,421 & 97,673 \\
    \textsc{sv} & \texttt{talbanken} & 4,303 & 504 & 1,219 & 6,026 & 967 & 23 & 196 & 1,186 & 66,646 & 9,797 & 20,377 & 96,820 \\
    \multirow{2}{*}{\textsc{zh}} & \texttt{gsd} & 3,997 & 500 & 500 & 4,997 & 6,136 & 754 & 767 & 7,657 & 98,616 & 12,663 & 12,012 & 123,291 \\
     & \texttt{gsdsimp} & 3,997 & 500 & 500 & 4,997 & 6,118 & 753 & 763 & 7,634 & 98,616 & 12,663 & 12,012 & 123,291 \\
     \midrule
     \textsc{de} & \texttt{pud} & -- & -- & 1,000 & 1,000 & -- & -- & 1,039 & 1,039 & -- & -- & 21,331 & 21,331 \\
     \textsc{en} & \texttt{pud} & -- & -- & 1,000 & 1,000 & -- & -- & 1,038 & 1,038 & -- & -- & 21,176 & 21,176 \\
     \textsc{pt} & \texttt{pud} & -- & -- & 1,000 & 1,000 & -- & -- & 1,099 & 1,099 & -- & -- & 23,407 & 23,407 \\
     \textsc{ru} & \texttt{pud} & -- & -- & 1,000 & 1,000 & -- & -- & 1,036 & 1,036 & -- & -- & 19,355 & 19,355 \\
     \textsc{sv} & \texttt{pud} & -- & -- & 1,000 & 1,000 & -- & -- & 1,029 & 1,029 & -- & -- & 19,076 & 19,076 \\
     \textsc{zh} & \texttt{pud} & -- & -- & 1,000 & 1,000 & -- & -- & 1,137 & 1,137 & -- & -- & 21,415 & 21,415 \\
     \midrule
      \textsc{ceb} & \texttt{gja} & -- & -- & 188 & 188 & -- & -- & 49 & 49 & -- & -- & 1,295 & 1,295 \\
      \multirow{2}{*}{\textsc{tl}} & \texttt{trg} & -- & -- & 128 & 128 & -- & -- & 92 & 92 & -- & -- & 734 & 734 \\
      & \texttt{ugnayan} & -- & -- & 94 & 94 & -- & -- & 61 & 61 & -- & -- & 1,097 & 1,097 \\
    \bottomrule
    \end{tabular}
    }
    \caption{Universal NER has broad coverage of named entities in several languages and domains, adding annotations to the development, testing, and training sets from Universal Dependencies \citep{nivre-etal-2020-universal}.}
    \label{tab:dataset-stats}
\end{table*}

\paragraph{Dataset Transfer}
Most of the included datasets are annotated from scratch using the annotation process detailed above, but a few (\textsc{da} \texttt{ddt}, \textsc{qaf} \texttt{arabizi}, \textsc{hr} and \textsc{sr} \texttt{set}) are transferred from other sources.
The Danish \texttt{ddt} annotations are derived from the \textit{News} portion of the DaN+ dataset~\cite{plank-etal-2020-DaN}; this text corresponds to the Universal Dependencies \texttt{ddt} treebank.
The Croatian \texttt{hr} annotations come from the \texttt{hr500k} dataset~\cite{ljubesic-etal-2016-new}, half of which, consisting of newspaper and various web texts, was used for producing the Croatian Universal Dependencies \texttt{hr\_set} treebank~\cite{agic-ljubesic-2015-universal}.
The NArabizi \texttt{arabizi} dataset was annotated on UD data using a slightly different NER schema and then automatically converted to the \shortname{} schema.
The Serbian \texttt{sr} data come from the \texttt{SETimes.SR} dataset~\cite{batanovic2018setimes}, which was used in its fullness to produce the Serbian Universal Dependencies \texttt{sr\_set} treebank~\cite{samardzic-etal-2017-universal}.
The original Croatian and Serbian NER annotations were annotated and curated in multiple iterations by various native speakers.
However, the annotations added to the \shortname{} dataset were slightly modified to conform to the \shortname{} annotation guidelines; namely, while nationalities and similar groups are annotated as \texttt{PER} in the original dataset, in the \shortname{} dataset such entities are omitted. Finally, we retain the original annotations from existing NER datasets in the ``xner'' label column.

\section{Universal NER: Statistics and Analysis}
\label{sec:stats}

This section presents an overview of the Universal NER (\shortname) dataset.
\shortname{}~v1 adds a NER annotation layer to 19 datasets (primarily treebanks from UD). It covers 13 geneologically and typologically diverse languages: Cebuano, Danish, German, English, Croatian, Narabizi, Portuguese, Russian, Slovak, Serbian, Swedish, Tagalog, and Chinese\footnote{Languages sorted by their ISO 639-1/639-2 codes \citep{ISO639-1,ISO639-2}}. Overall, \shortname{}~v1 contains ten full datasets with training, development, and test splits over nine languages, three evaluation sets for lower-resource languages (\textsc{tl} and \textsc{ceb}), and a parallel evaluation benchmark spanning six languages.

\subsection{Dataset Statistics}
In \autoref{tab:dataset-stats}, we report the number of sentences, tokens, and annotated entities for each dataset in \shortname{}.
The datasets in \shortname{} cover a wide range of data quantities: some provide a limited amount of evaluation data for a commonly low-resourced language, whereas others annotate thousands of training and evaluation sentences.

\begin{table*}[htbp]
\centering
\scalebox{0.8}{
\begin{tabular}{ll rrrr rrrr rrrr}
\toprule
& & \multicolumn{4}{c}{Train} & \multicolumn{4}{c}{Dev} & \multicolumn{4}{c}{Test} \\
\cmidrule(lr){3-6} \cmidrule(lr){7-10} \cmidrule(lr){11-14}
Lang. & Dataset & \texttt{LOC} & \texttt{ORG} & \texttt{PER} & \% Docs & \texttt{LOC} & \texttt{ORG} & \texttt{PER} & \% Docs & \texttt{LOC} & \texttt{ORG} & \texttt{PER} & \% Docs \\
\midrule
\textsc{da} & \texttt{ddt} & .875 & .778 & .959 & 100\% & .917 & .765 & .934 & 100\% & .882 & .805 & .975 & 100\% \\
\textsc{en} & \texttt{ewt} & .696 & .533 & .925 & 20\% & .786 & .640 & .949 & 20\% & .825 & .869 & .969 & 20\% \\
\textsc{pt} & \texttt{bosque} & .928 & .902 & .974 & 11\% & .850 & .885 & .980 & 25\% & .955 & .914 &.975 & 23\% \\
\textsc{sk} & \texttt{snk} & .840 & .743 & .900 & 100\% & .801 & .597 & .770 & 100\% & .837 & .621 & .823 & 100\% \\
\textsc{sv} & \texttt{talbanken} & .857 & .670 & .913 & 100\% & .800 & .461 & .888 & 100\% & .937 & .812 & .871 & 100\% \\
\textsc{zh} & \texttt{gsd} & .800 & .724 & .917 & 14\% & .795 & .661 & .956 & 100\% & .860 & .711  & .944 & 23\% \\
\midrule
\textsc{de} & \texttt{pud} & -- & -- & -- & -- & -- & -- & -- & -- & .709 & .840 & .812 & 6\% \\
\textsc{en} & \texttt{pud} & -- & -- & -- & -- & -- & -- & -- & -- & 1.00 & .936 & .966 & 6\% \\
\textsc{pt} & \texttt{pud} & -- & -- & -- & -- & -- & -- & -- & -- & .903 & .920 & .985 & 14\% \\
\textsc{ru} & \texttt{pud} & -- & -- & -- & -- & -- & -- & -- & -- & .719 & .531 & .891 & 100\% \\
\textsc{sv} & \texttt{pud} & -- & -- & -- & -- & -- & -- & -- & -- & .865 & .735 & .944 & 100\% \\
\textsc{zh} & \texttt{pud} & -- & -- & -- & -- & -- & -- & -- & -- & .752 & .776 & .971 & 20\% \\
\midrule
\textsc{ceb} & \texttt{gja} & -- & -- & -- & -- & -- & -- & -- & -- & .769 & 1.00 & .914 & 71\% \\
\multirow{2}{*}{\textsc{tl}} & \texttt{trg} & -- & -- & -- & -- & -- & -- & -- & -- & .833 & -- & .957 & 100\% \\
 & \texttt{ugnayan} & -- & -- & -- & -- & -- & -- & -- & -- & .913 & -- & -- & 100\% \\

\bottomrule
\end{tabular}
}

\caption{Inter-annotator agreement scores for the datasets annotated natively for the Universal NER project. We don't report IAA for the datasets adapted from other sources, or from \texttt{zh\_gsdsimp}, which has nearly identical annotations to \texttt{zh\_gsd}. \textbf{\% Docs} refers to the percentage of documents annotated by multiple annotators.}
\label{tab:iaa-scores}
\end{table*}

\begin{figure}[b!]
    \centering
    \includegraphics[width=0.8\linewidth]{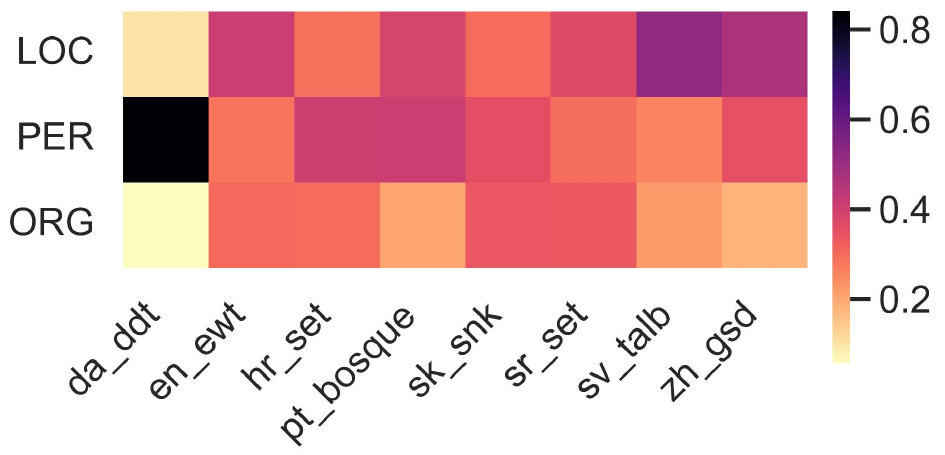}
    \caption{Distribution of tags in different \shortname{} training sets. \texttt{zh\_gsdsimp} has the same distribution as \texttt{zh\_gsd}.}
    \label{fig:train-dist}
\end{figure}

The datasets in \shortname{} also cover a diverse range of domains, spanning web sources such as social media to more traditional provenances like news text.
\autoref{tab:uner-domains} in the appendix presents the complete set of sources for the data and the distribution of NER tags in each dataset, along with references to each original treebank paper.
The variety in data sources leads to varied distributions of tags across datasets (\autoref{fig:train-dist}).

\subsection{Inter-Annotator Agreement}
\label{sec:iaa}
We calculate inter-annotator agreement (IAA, \autoref{tab:iaa-scores}) for each dataset in \shortname{} that was annotated with the above process and for which we have secondary annotations. %
Table~\ref{tab:iaa-scores} reports agreement as per-label F\textsubscript{1} score, using one annotator as ``reference,'' and the other as ``prediction.''

\paragraph{\texttt{ORG} vs \texttt{LOC} Confusion}
\label{ssec:olconfusion}
The agreement on \texttt{ORG} and \texttt{LOC} is generally lower than that on \texttt{PER}. The annotation guidelines allow certain named entities to take either the \texttt{ORG} or \texttt{LOC} tag based on context. In some cases, the context is underspecified, leading to ambiguity. For example, a restaurant is a \texttt{LOC} when you go there to eat, but it is an \texttt{ORG} when it hires a new chef. A city is a \texttt{LOC} when you move there, but it is an \texttt{ORG} when it levies taxes. Officially, it is the \textit{city government} that levies taxes, but common usage allows, for example, ``\nerexample{Springfield}{ORG} charges a brutal income tax.'' 
CoNLL 2003 English also has this ambiguity, with many documents where city names, representing sports teams, are annotated as \texttt{ORG}.
We find this ambiguity is particularly common in the \texttt{en\_ewt} train and validation splits, primarily in documents in the \textit{reviews} domain, which are short and very informal (e.g. ``we love pamelas'').

\subsection{Agreement with the \texttt{PROPN} POS Tag}
\label{ssection:propn}
The proper noun (\texttt{PROPN}) part-of-speech tag used in UD represents the subset of nouns that are used as the name of a specific person, place, or object \cite{nivre-etal-2020-universal}.
We hypothesize that named entities as defined in \shortname{} act roughly as a subset of these \texttt{PROPN} words or phrases, although not a strict subset due to divergent definitions.
To test this, we calculate the precision of the \shortname{} annotations against the UD \texttt{PROPN} tags (\autoref{tab:propn_overlap}, F\textsubscript{1} scores reported in \autoref{tab:propn_overlap_f1}).
Overall, precision is relatively high, with a mean precision of 0.761 across datasets.
Lower precision is often due to multi-word names containing non-\texttt{PROPN} words (e.g.,~``Catherine the Great'').
The differences in precision can also be due to language-specific \texttt{PROPN} annotation guidelines: for example, while the English PUD treebank tags the United States entity as ``\nerexample{United}{PROPN} \nerexample{States}{PROPN}'', Russian PUD tags it as {\fontencoding{T2A}\selectfont ``\nerexample{Соединенных}{{\fontencoding{T1}\selectfont ADJ}} \nerexample{Штатов}{{\fontencoding{T1}\selectfont NOUN}}''}.

\begin{table}[t!]
\centering
\scalebox{0.8}{
\begin{tabular}{llccc}
\toprule
Lang. & Dataset & Train & Dev & Test \\
\midrule
\textsc{da} & \texttt{ddt} & .709 & .729 & .722 \\
\textsc{en} & \texttt{ewt} & .890 & .895 & .892 \\
\textsc{hr} & \texttt{set} & .683 & .651 & .671 \\
\textsc{pt} & \texttt{bosque} & .864 & .881 & .844 \\
 \textsc{qaf} & \texttt{arabizi} & .952 & .960 & .985 \\
\textsc{sk} & \texttt{snk} & .803 & .783 & .688 \\
\textsc{sr} & \texttt{set} & .687 & .631 & .680 \\
\textsc{sv} & \texttt{talbanken} & .766 & .756 & .842  \\
\textsc{zh} & \texttt{gsd} & .605 & .624 & .616 \\
\textsc{zh} & \texttt{gsdsimp} & .601 & .604 & .617 \\
\midrule
\textsc{de} & \texttt{pud} & -- & -- & .712 \\
\textsc{en} & \texttt{pud} & -- & -- & .872 \\
\textsc{pt} & \texttt{pud} & -- & -- & .749 \\
\textsc{ru} & \texttt{pud} & -- & -- & .708 \\
\textsc{sv} & \texttt{pud} & -- & -- & .810 \\
\textsc{zh} & \texttt{pud} & -- & -- & .634 \\
\midrule
\textsc{ceb} & \texttt{gja} & -- & -- & .980 \\
\textsc{tl} & \texttt{trg} & -- & -- & .958 \\
\textsc{tl} & \texttt{ugnayan} & -- & -- & .654 \\

\bottomrule
\end{tabular}%
}
\caption{Comparing the overlap (Precision) between \shortname{} annotations and UD PROPN tags.}
\label{tab:propn_overlap}
\end{table}

\begin{figure*}[ht]
\centering
\begin{subfigure}{0.32\textwidth}
    \includegraphics[width=\textwidth]{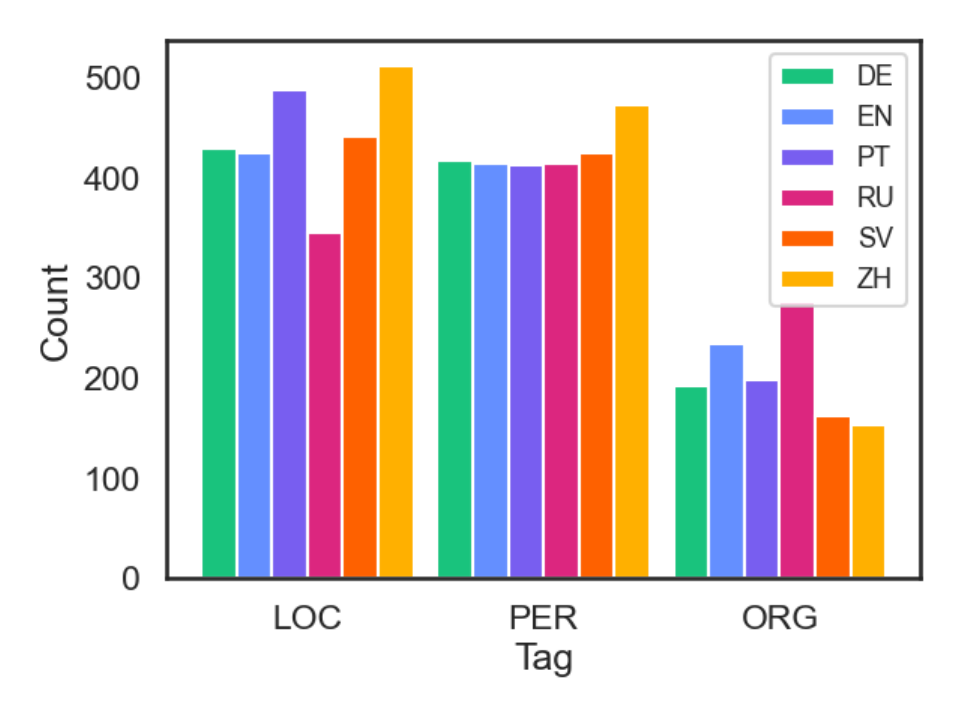}
\end{subfigure}
\hfill
\begin{subfigure}{0.32\textwidth}
    \includegraphics[width=\textwidth]{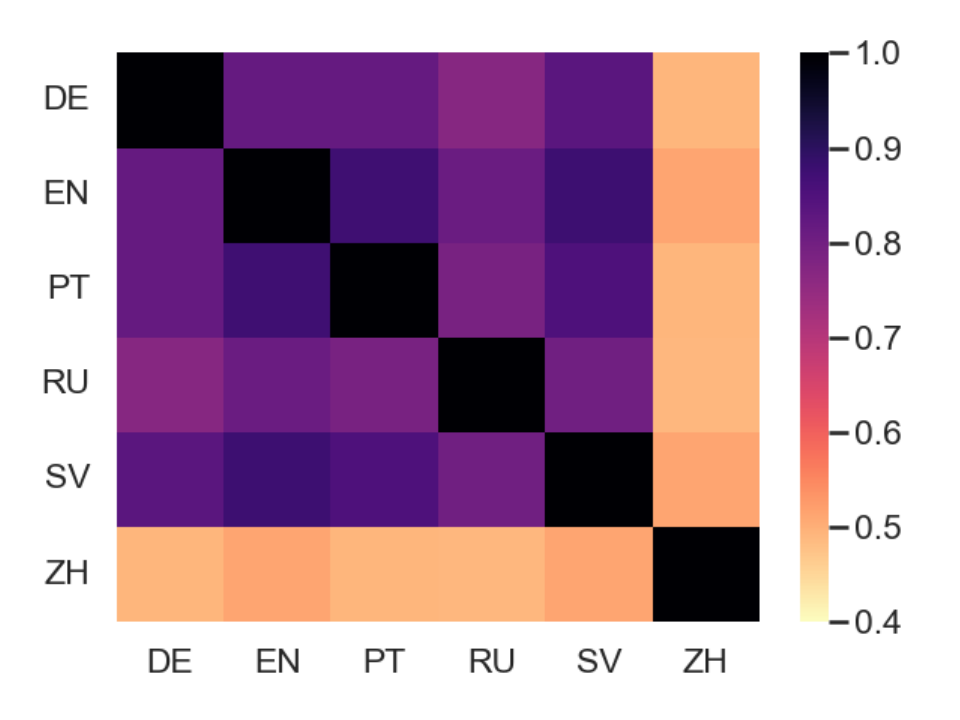}
\end{subfigure}
\hfill
\begin{subfigure}{0.32\textwidth}
    \includegraphics[width=\textwidth]{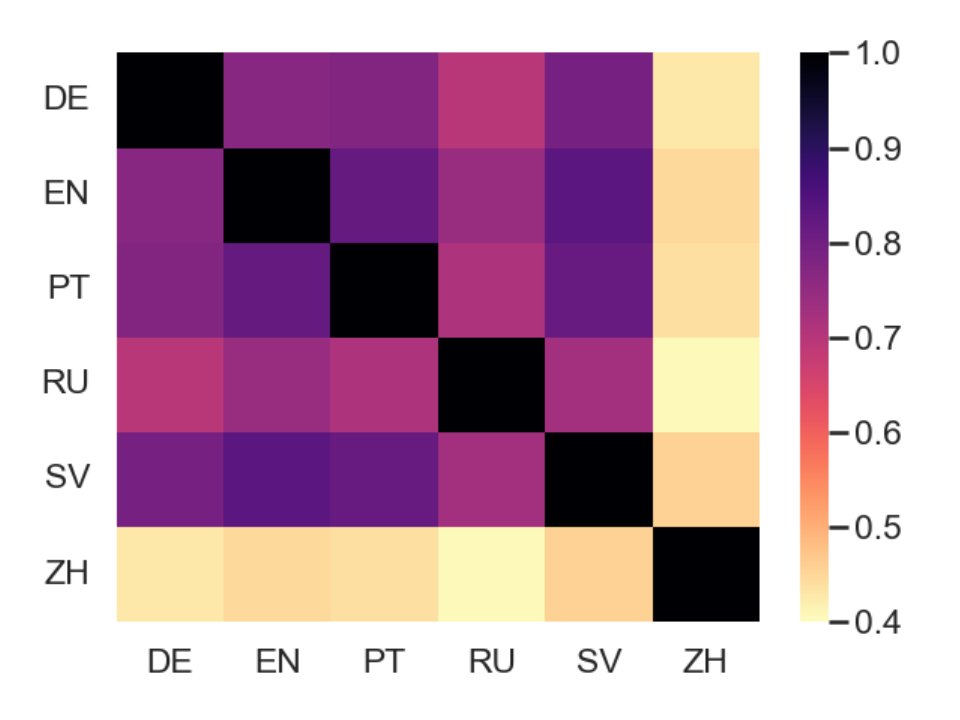}
\end{subfigure}
\caption{Cross-lingual comparison of NER Annotations on top of PUD treebanks. \textbf{Left}: Global distribution of tags for each PUD language. \textbf{Center}: Sentence-level agreement between languages for the number of entities. \textbf{Right}: Sentence-level agreement between languages for the identity of entities.}
\label{fig:pud-comp}
\end{figure*}

\subsection{Cross-lingual Agreement in \shortname{}}
\label{ssection:pud-agreement}
\shortname{} contains sentence-aligned evaluation sets for six languages (German, English, Portuguese, Russian, Swedish, and Chinese) that are annotated on top of the Parallel Universal Dependencies treebanks~\cite[PUD; ][]{zeman-etal-2017-conll}. \autoref{fig:pud-comp} summarizes the similarity of the NER annotations across these target languages in PUD.

We find that the overall distribution of tags is similar for the Western European languages (left panel): the English, German, and Swedish annotations contain very similar counts of \texttt{LOC} and \texttt{PER} entities, with slightly more variance in \texttt{ORG} tags. Portuguese has a similar distribution with slightly more \texttt{LOC} entities. However, the Russian and Chinese annotations contain differing distributions from both these languages and each other. 

A similar trend occurs in the sentence-level pairwise agreement on entity counts and identities between languages (center). There is relatively high agreement on the number of entities between European languages, with Russian differing slightly more from English, German, Portuguese, and Swedish. However, the Chinese benchmark agrees less frequently: the Chinese annotations match other languages on the number of entities in 50.4\% of sentences; the other languages have an average agreement of 71.7--75.6\%. Pairwise agreement on the specific entities in a given sentence shows similar behavior, albeit with lower agreement overall (right). 

Many of these annotation differences likely stem from the translation process. While the data is aligned at the sentence level, linguistic variation and translator decisions may cause an entity to be added to or removed from the sentence, or the concept may be expressed in a manner that no longer qualifies as a named entity under the annotation guidelines.\footnote{Consider the phrases:  \zh{“奧巴馬對在北卡羅來納大學運動場上的群眾說道。”} and ``he told the crowd gathered on a sports field at the University of North Carolina.'' In Chinese, \textit{Obama} (\zh{奧巴馬}) is referred to by name, whereas the English version uses a pronoun.} %
While we cannot directly measure inter-annotator agreement across languages because of the above differences, some variation also undoubtedly stems from annotation differences and errors, just as these cause disagreement between annotators on the same benchmark.

\begin{figure*}[t]
\centering
\includegraphics[width=0.9\textwidth]{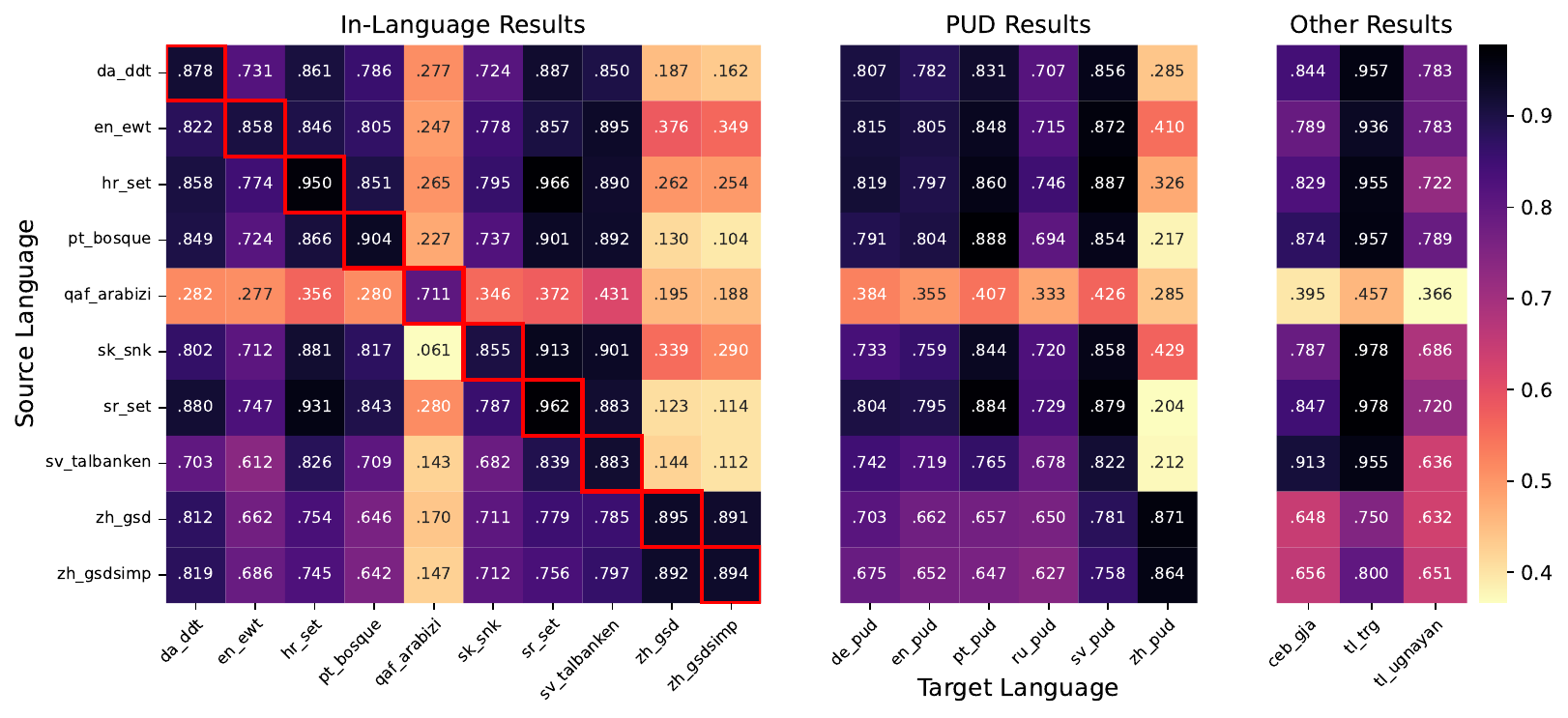}
\caption{Heatmap of micro F\textsubscript{1} scores on test sets with different fine-tuned models. The y-axis indicates the dataset that the model is fine-tuned on, and the x-axis indicates the datasets that the models are evaluated on. \textbf{Left}: Model performance on datasets that contains the train, dev and test splits. The highlighted diagonal cells are the in-dataset results. \textbf{Center}: Model performance on the PUD datasets. \textbf{Right}: Model performance on all other datasets.}
\label{fig:baseline}
\end{figure*}

In the case of Chinese and English, we manually audited the annotation discrepancies. The differences in the \texttt{LOC} and \texttt{ORG} tags mainly stem from the confusion outlined in Section~\ref{ssec:olconfusion}.
Additionally, we observed more than 30 instances that could be explained by language-specific morphological inflection rules.
Specifically, country names are used directly to modify the following nouns in Chinese as opposed to English using the adjectival form.\footnote{I.e., \zh{“韓國公司”} `South Korean company'. The Chinese word \zh{“韓國”} means the country `South Korea', and in this case, directly modifies the noun \zh{“公司”} `company'. This word was consequently labeled as \texttt{LOC}, whereas its English counterpart is \texttt{O}.}
Finally, the increase in \texttt{PER} entities can be best explained by the style of Chinese writing, which tends to transliterate non-Chinese names into Chinese and append the Latin name in parentheses; in these cases, each instance of the name would be tagged as a separate \texttt{PER} entity.\footnote{An example is ``\zh{聖羅斯季斯拉夫} (St. Rastislav)'', in which the English name is parenthesized and kept in the Chinese sentence, causing both names to be annotated.}

\section{Baselines for \shortname{}}
\label{sec:baselines}
This section establishes initial baselines on the datasets in \shortname{}~v1 and provides in-language and cross-lingual results with XLM-R\textsubscript{Large}.

\subsection{Experiment Setup}
We finetune XLM-R\textsubscript{Large} (560M parameters)~\citep{conneau-etal-2020-unsupervised} on the UNER datasets in which train and dev sets are available,\footnote{\texttt{ddt}, \texttt{ewt}, \texttt{set}, \texttt{bosque}, \texttt{arabizi}, \texttt{snk}, \texttt{set}, \texttt{talbanken}, \texttt{gsd}, \texttt{gsdsimp}} using a single NVIDIA GeForce RTX 3090 GPU.
We also evaluate the performance of XLM-R\textsubscript{Large} jointly finetuned on all training sets (\texttt{all}) listed above. We use a learning rate of 3e-5 and batch size of 16, except for \texttt{bosque}, where we used a batch size of 8, and batch size of 4 in the cases of \texttt{talbanken} and \texttt{all}. All the code we used is adapted from the Huggingface \texttt{transformers} package~\citep{wolf-etal-2020-transformers}.

\subsection{Results and Discussion}
\autoref{fig:baseline} reports the micro F\textsubscript{1} scores on all test sets when XLM-R\textsubscript{Large} is finetuned on different languages. The in-language performance shown on the diagonal on the left of \autoref{fig:baseline} is almost always the highest among all test sets, with a few exceptions such as Simplified Chinese vs Traditional Chinese (\textsc{zh}) and Croatian (\textsc{hr}) vs Serbian (\textsc{sr}). This most likely stems from the fact that both pairs are closely related languages.

We also observe that in most cases (i.e., between European languages), cross-lingual transfer performs well, achieving over .600 F\textsubscript{1}. However, transfer results in strikingly low performance on all three Chinese datasets \{\texttt{gsd}, \texttt{gsdsimp}, \texttt{pud}\}, as well as on the Maghrebi-Arabic-French (\textsc{QAF}) dataset \{\texttt{arabizi}\}.
The results on the Chinese datasets align with observations from previous work~\citep{chen-etal-2023-frustratingly,wu-etal-2020-single,bao-etal-2019-low} that other languages do not transfer well to Chinese.
Narabizi is a North-African Arabic dialect written in Latin script that often involves code-switching with French. The lack of similarities between this language and all other languages in our dataset might have resulted in poor transfer performance. Furthermore, Narabizi --- along with Cebuano --- are not included in the pretraining languages for XLM-R, which likely also affects their performance in this setting.

\autoref{tab:baseline_results} (in the Appendix) shows the tag-level performance breakdown. For all languages, F\textsubscript{1} on \texttt{ORG} is always the lowest, and \texttt{LOC} is almost always the second lowest.
This likely stems from the similarity between \texttt{ORG} and \texttt{LOC} entities discussed in Section~\ref{ssec:olconfusion}, whereas the names of people are usually less ambiguous, resulting in the highest F\textsubscript{1} on \texttt{PER} for most datasets. Overall, the trained models finetuned on the UNER datasets exhibit promising results, and we leave further improvements on multi- and cross-lingual NER with these datasets to future work.

\begin{figure}
    \centering
    \includegraphics[width=\linewidth]{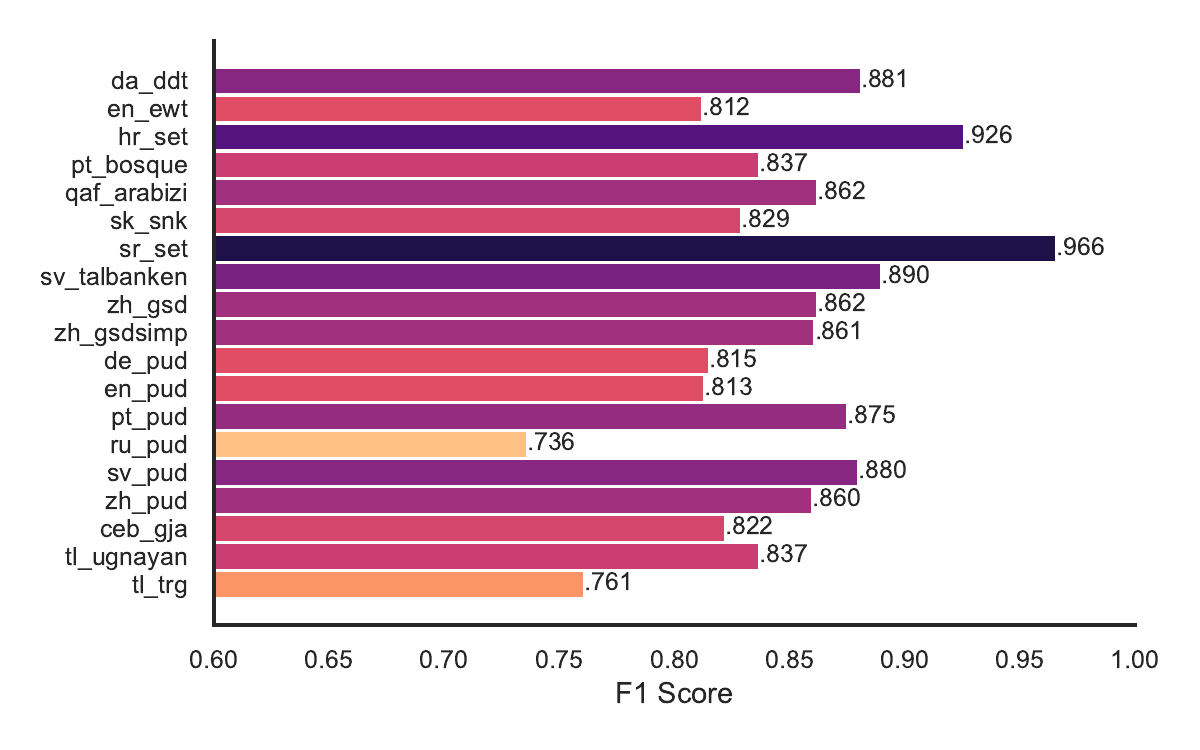}
    \vspace{-0.8cm} %
    \caption{F\textsubscript{1} scores of each UNER test set after finetuning XLM-R\textsubscript{Large} on \texttt{all} training sets.}
    \label{fig:all-accuracy}
\end{figure}

Finally, the performance of the model finetuned on \texttt{all} is included in \autoref{fig:all-accuracy}.
Most \texttt{all} F\textsubscript{1} scores are similar to the F\textsubscript{1} scores from individual training sets or lead to a moderate decrease in performance; however, in some cross-lingual cases the joint training improves performance, such as on \texttt{zh\_pud} which improved from .410 using a model finetuned on \texttt{en\_ewt} to .860.
Finetuning on a diverse multilingual dataset helps preserve and even improve the performance on benchmarks in diverse languages.

\section{Related Work}
\label{sec:rw}

\paragraph{Adding a NER layer to UD}
Some single-language efforts have added a manually annotated NER layer to emerging or existing UD data. \citet{agic-ljubesic-2014-setimes} annotated the \texttt{SETimes.HR} dataset with linguistic and NER information, becoming the \texttt{set\_hr} UD dataset later~\cite{agic-ljubesic-2015-universal}. \citet{plank-2019-neural} added a layer of NER to the dev and test portions of the Danish UD treebank (DDT) for cross-lingual evaluation; \citet{plank-etal-2020-DaN} fully annotated it with nested NER entities.
\citet{hvingelby-etal-2020-dane} annotated the same Danish UD data with a flat annotation scheme.

Other languages have seen efforts in a similar spirit. \citet{jorgensen-etal-2020-norne} added a named entity annotation layer on top of the Norwegian Dependency Treebank, \citet{luoma-etal-2020-broad} built the Turku NER corpus, and \citet{plank-2021-cross} added a layer on top of English EWT. Recently, \citet{muischnek-muurisep-2023-named} introduced the largest publicly available Estonian NER dataset. Complementing these efforts, \citet{riabi-etal-2023-enriching} added several annotation layers, including NER, to the NArabizi treebank \cite{seddah-etal-2020-building}, a North-African Arabic dialect dataset written in Latin script with a high-level of language variability and code-switching.

\paragraph{Multilingual NER resources}

Several benchmark datasets for NER offer coverage for a variety of representative languages.
Aside from well-known benchmarks such as CoNLL 2002/2003~\citep{tjong-kim-sang-2002-introduction,tjong-kim-sang-de-meulder-2003-introduction}, other datasets were built to address a unique need, such as focusing on low-resource languages like LORELEI \citep{strassel-tracey-2016-lorelei} or incorporating particularly challenging annotations, as seen in MultiCoNER \citep{malmasi-etal-2022-multiconer,malmasi-etal-2022-semeval}.
MasakhaNER~\citep{adelani-etal-2022-masakhaner} harnessed the \textit{Masakhane} community to produce gold-standard annotations for ten African languages.

Other datasets for multilingual and non-English NER use a silver-standard annotation process ~\citep{nothman2013learning,pan-etal-2017-cross,tedeschi-etal-2021-wikineural-combined}. Nonetheless, CoNLL 2002/2003 remains one of the main benchmarks in multilingual NER.
A recent work, also called UNER~\citep{alves2020uner}, attempts to produce silver-standard corpora by propagating English annotations across parallel corpora but with no baseline evaluations.
Lastly, another contemporary work called Universal NER~\citep{zhou2023universalner} bears no relation to our effort as it contains no annotation component.

\paragraph{Modeling for multilingual NER}

Several works have explored the task of NER outside of English.
The earliest build language-independent methods ~\citep[\textit{inter alia}]{cucerzan-yarowsky-1999-language, lample-etal-2016-neural}.
Cross-lingual techniques have also emerged to transfer information between languages, especially from high- to low-resource languages~\citep{ruder-etal-2019-unsupervised} or combining model and data transfer across languages~\citep{wu-etal-2020-unitrans}.
Currently, the standard paradigm for multilingual NER involves finetuning or prompting multilingual language models~\citep[e.g.,][]{wu-dredze-2020-languages, muennighoff-etal-2023-crosslingual}. \shortname{} supports these modeling efforts by providing gold-standard annotations across various languages.

\paragraph{Community-driven annotation projects} 
The field of NLP has been shaped by community-driven annotation projects. 
One prime example is the Universal Dependencies (UD) project~\citep{nivre-etal-2020-universal}, precipitated by the earlier introduction of the universal POS tagset~\citep{mcdonald-etal-2013-universal}.
Extensions and sister projects to UD have emerged~\citep[e.g.,][]{savary-etal-2023-parseme,kahane-etal-2021-annotation}, to which \shortname{} is now added.
Another notable endeavor is UniMorph~\citep{kirov-etal-2018-unimorph,mccarthy-etal-2020-unimorph}, which covers 182 languages~\citep{batsuren-etal-2022-unimorph,batsuren-etal-2021-morphynet}.
The Masakhane Project has also produced several high-quality community efforts \cite{adelani-etal-2021-masakhaner,adelani-etal-2022-masakhaner,dione-etal-2023-masakhapos,dione2023masakhapos}.

The \shortname{} project follows the same community-driven approach by asking volunteers to contribute annotations for their respective languages.

\section{Conclusion}
\label{sec:conclusion}

We introduce Universal NER (\shortname), a gold-standard data initiative covering 13 languages for named entity recognition (NER). The datasets included in \shortname{} v1 cover a wide variety of domains and language families, and we establish initial performance metrics for these benchmarks. \shortname{} opens several opportunities for research in NER outside of English and for cross-lingual transfer; in particular, this project provides human-annotated and standardized evaluations for multilingual NER. 

After releasing the current version of the \shortname{} project, we plan to expand language coverage and diversity of this effort by both recruiting additional annotators and integrating existing NER datasets when possible. This will also allow us to obtain more robust agreement measures and verify the quality of existing annotations in UNER. In the longer term, our aims for Universal NER include rigorous quality checking of annotation results for robustness and further integration of finetuned models and data analysis tools into the project.

\section*{Limitations}

\paragraph{Dataset Domains and Languages}
The data included in UNER v1 covers a range of domains and languages, depending on the available annotators and datasets in UD (Appendix \autoref{tab:uner-domains}). The variance in domains and languages will generally affect the efficacy of cross-lingual learning and evaluation. However, we also provide a standardized, parallel evaluation set for a subset of the languages in \shortname{}. Furthermore, we invite researchers who would like to see additional languages in UNER to join the annotation effort. 

\paragraph{Springboarding from Universal Dependencies}
Our preliminary criterion for languages and data to be included in the current version of \shortname{} is that it should be already in the Universal Dependencies (UD)~\citep{de-marneffe-etal-2021-universal}.
This is to ensure the quality of the underlying data and to facilitate research in conjunction with existing UD treebanks, which include part-of-speech tags, tokenization, lemmas, and glosses. However, future iterations of the \shortname{} initiative are open to all languages, especially low-resource ones, regardless of whether they are present in UD.

\paragraph{Number of Annotators}
The \shortname{} project relies on crowd-sourcing and community participation for annotation efforts. Thus, the languages included have varying numbers of annotators who have accepted the invitation to contribute. Nonetheless, as reported in \autoref{tab:iaa-scores}, each language has at least \textbf{two} annotators for a subset of its documents and thus a corresponding measure of inter-annotator agreement.

\section*{Ethics Statement}
Our annotated data is built on top of Universal Dependencies, an already established data resource. Thus, we do not foresee any serious or harmful issues arising from its content.
Interested volunteer annotators who were invited to the project have also been informed of the guidelines as discussed in Section~\ref{sec:annotation} for annotating NER-ready datasets before starting with the process.

\section*{Acknowledgments}
This project could not have happened without the enthusiastic response and hard work of many annotators in the NLP community, and for that we are extremely grateful.
Annotators additional to authors are: Elyanah Aco, Ekaterina Artemova, Vuk Batanović, Jay Rhald Caballes Padilla, Chunyuan Deng, Ivo-Pavao Jazbec, Juliane Karlsson, Jozef Kubík, Peter Krantz, Myron Darrel Montefalcon, Stefan Schweter, Sif Sonniks, Emil Stenström, Miriam Šuppová.

We would like to thank Joakim Nivre, Dan Zeman, Matthew Honnibal, Željko Agić, Constantine Lignos, and Amir Zeldes for early discussion and helpful ideas at the very beginning of this project.

JMI is funded by National University Philippines and the UKRI Centre for Doctoral Training in Accountable, Responsible and Transparent AI [EP/S023437/1] of the University of Bath.

Arij Riabi is funded by the European Union’s Horizon 2020
research and innovation programme under grant
agreement No. 101021607. 

Marek \v{S}uppa was partially supported by the grant APVV-21-0114.

\bibliography{anthology,uner}

\begin{thebibliography}{69}
\expandafter\ifx\csname natexlab\endcsname\relax\def\natexlab#1{#1}\fi

\bibitem[{Adelani et~al.(2022)Adelani, Neubig, Ruder, Rijhwani, Beukman, Palen-Michel, Lignos, Alabi, Muhammad, Nabende, Dione, Bukula, Mabuya, Dossou, Sibanda, Buzaaba, Mukiibi, Kalipe, Mbaye, Taylor, Kabore, Emezue, Aremu, Ogayo, Gitau, Munkoh-Buabeng, Memdjokam~Koagne, Tapo, Macucwa, Marivate, Elvis, Gwadabe, Adewumi, Ahia, Nakatumba-Nabende, Mokono, Ezeani, Chukwuneke, Oluwaseun~Adeyemi, Hacheme, Abdulmumin, Ogundepo, Yousuf, Moteu, and Klakow}]{adelani-etal-2022-masakhaner}
David Adelani, Graham Neubig, Sebastian Ruder, Shruti Rijhwani, Michael Beukman, Chester Palen-Michel, Constantine Lignos, Jesujoba Alabi, Shamsuddeen Muhammad, Peter Nabende, Cheikh M.~Bamba Dione, Andiswa Bukula, Rooweither Mabuya, Bonaventure F.~P. Dossou, Blessing Sibanda, Happy Buzaaba, Jonathan Mukiibi, Godson Kalipe, Derguene Mbaye, Amelia Taylor, Fatoumata Kabore, Chris~Chinenye Emezue, Anuoluwapo Aremu, Perez Ogayo, Catherine Gitau, Edwin Munkoh-Buabeng, Victoire Memdjokam~Koagne, Allahsera~Auguste Tapo, Tebogo Macucwa, Vukosi Marivate, Mboning~Tchiaze Elvis, Tajuddeen Gwadabe, Tosin Adewumi, Orevaoghene Ahia, Joyce Nakatumba-Nabende, Neo~Lerato Mokono, Ignatius Ezeani, Chiamaka Chukwuneke, Mofetoluwa Oluwaseun~Adeyemi, Gilles~Quentin Hacheme, Idris Abdulmumin, Odunayo Ogundepo, Oreen Yousuf, Tatiana Moteu, and Dietrich Klakow. 2022.
\newblock \href {https://doi.org/10.18653/v1/2022.emnlp-main.298} {{M}asakha{NER} 2.0: {A}frica-centric transfer learning for named entity recognition}.
\newblock In \emph{Proceedings of the 2022 Conference on Empirical Methods in Natural Language Processing}, pages 4488--4508, Abu Dhabi, United Arab Emirates. Association for Computational Linguistics.

\bibitem[{Adelani et~al.(2021)Adelani, Abbott, Neubig, D{'}souza, Kreutzer, Lignos, Palen-Michel, Buzaaba, Rijhwani, Ruder, Mayhew, Azime, Muhammad, Emezue, Nakatumba-Nabende, Ogayo, Anuoluwapo, Gitau, Mbaye, Alabi, Yimam, Gwadabe, Ezeani, Niyongabo, Mukiibi, Otiende, Orife, David, Ngom, Adewumi, Rayson, Adeyemi, Muriuki, Anebi, Chukwuneke, Odu, Wairagala, Oyerinde, Siro, Bateesa, Oloyede, Wambui, Akinode, Nabagereka, Katusiime, Awokoya, MBOUP, Gebreyohannes, Tilaye, Nwaike, Wolde, Faye, Sibanda, Ahia, Dossou, Ogueji, DIOP, Diallo, Akinfaderin, Marengereke, and Osei}]{adelani-etal-2021-masakhaner}
David~Ifeoluwa Adelani, Jade Abbott, Graham Neubig, Daniel D{'}souza, Julia Kreutzer, Constantine Lignos, Chester Palen-Michel, Happy Buzaaba, Shruti Rijhwani, Sebastian Ruder, Stephen Mayhew, Israel~Abebe Azime, Shamsuddeen~H. Muhammad, Chris~Chinenye Emezue, Joyce Nakatumba-Nabende, Perez Ogayo, Aremu Anuoluwapo, Catherine Gitau, Derguene Mbaye, Jesujoba Alabi, Seid~Muhie Yimam, Tajuddeen~Rabiu Gwadabe, Ignatius Ezeani, Rubungo~Andre Niyongabo, Jonathan Mukiibi, Verrah Otiende, Iroro Orife, Davis David, Samba Ngom, Tosin Adewumi, Paul Rayson, Mofetoluwa Adeyemi, Gerald Muriuki, Emmanuel Anebi, Chiamaka Chukwuneke, Nkiruka Odu, Eric~Peter Wairagala, Samuel Oyerinde, Clemencia Siro, Tobius~Saul Bateesa, Temilola Oloyede, Yvonne Wambui, Victor Akinode, Deborah Nabagereka, Maurice Katusiime, Ayodele Awokoya, Mouhamadane MBOUP, Dibora Gebreyohannes, Henok Tilaye, Kelechi Nwaike, Degaga Wolde, Abdoulaye Faye, Blessing Sibanda, Orevaoghene Ahia, Bonaventure F.~P. Dossou, Kelechi Ogueji, Thierno~Ibrahima DIOP,
  Abdoulaye Diallo, Adewale Akinfaderin, Tendai Marengereke, and Salomey Osei. 2021.
\newblock \href {https://doi.org/10.1162/tacl_a_00416} {{M}asakha{NER}: Named entity recognition for {A}frican languages}.
\newblock \emph{Transactions of the Association for Computational Linguistics}, 9:1116--1131.

\bibitem[{Agi{\'c} and Ljube{\v{s}}i{\'c}(2014)}]{agic-ljubesic-2014-setimes}
{\v{Z}}eljko Agi{\'c} and Nikola Ljube{\v{s}}i{\'c}. 2014.
\newblock \href {http://www.lrec-conf.org/proceedings/lrec2014/pdf/690_Paper.pdf} {The {SET}imes.{HR} linguistically annotated corpus of {C}roatian}.
\newblock In \emph{Proceedings of the Ninth International Conference on Language Resources and Evaluation ({LREC}'14)}, pages 1724--1727, Reykjavik, Iceland. European Language Resources Association (ELRA).

\bibitem[{Agi{\'c} and Ljube{\v{s}}i{\'c}(2015)}]{agic-ljubesic-2015-universal}
{\v{Z}}eljko Agi{\'c} and Nikola Ljube{\v{s}}i{\'c}. 2015.
\newblock \href {https://aclanthology.org/W15-5301} {{U}niversal {D}ependencies for {C}roatian (that work for {S}erbian, too)}.
\newblock In \emph{The 5th Workshop on {B}alto-{S}lavic Natural Language Processing}, pages 1--8, Hissar, Bulgaria. INCOMA Ltd. Shoumen, BULGARIA.

\bibitem[{Alves et~al.(2020)Alves, Kuculo, Amaral, Thakkar, and Tadic}]{alves2020uner}
Diego Alves, Tin Kuculo, Gabriel Amaral, Gaurish Thakkar, and Marko Tadic. 2020.
\newblock \href {https://arxiv.org/pdf/2010.12406} {{UNER: Universal Named-Entity Recognition Framework}}.
\newblock \emph{arXiv preprint arXiv:2010.12406}.

\bibitem[{Aquino et~al.(2020)Aquino, de~Leon, and Bacolod}]{ud_tagalog_ugnayan}
Angelina Aquino, Franz de~Leon, and Mary~Ann Bacolod. 2020.
\newblock {UD\_Tagalog-Ugnayan}.
\newblock \url{https://github.com/UniversalDependencies/UD_Tagalog-Ugnayan}.

\bibitem[{Aranes(2022)}]{aranes-2022-gja}
Glyd Aranes. 2022.
\newblock \href {https://erepo.uef.fi/bitstream/handle/123456789/27525/urn_nbn_fi_uef-20220433.pdf?sequence=1} {{The GJA Cebuano Treebank: Creating a Cebuano Universal Dependencies Treebank}}.
\newblock Master's thesis, It{\"a}-Suomen yliopisto.

\bibitem[{Bao et~al.(2019)Bao, Huang, Li, and Zhu}]{bao-etal-2019-low}
Zuyi Bao, Rui Huang, Chen Li, and Kenny Zhu. 2019.
\newblock \href {https://doi.org/10.18653/v1/D19-1095} {Low-resource sequence labeling via unsupervised multilingual contextualized representations}.
\newblock In \emph{Proceedings of the 2019 Conference on Empirical Methods in Natural Language Processing and the 9th International Joint Conference on Natural Language Processing (EMNLP-IJCNLP)}, pages 1028--1039, Hong Kong, China. Association for Computational Linguistics.

\bibitem[{Batanovi\'{c} et~al.(2018)Batanovi\'{c}, Ljube{\v{s}}i\'{c}, and Samard\v{z}i\'{c}}]{batanovic2018setimes}
Vuk Batanovi\'{c}, Nikola Ljube{\v{s}}i\'{c}, and Tanja Samard\v{z}i\'{c}. 2018.
\newblock {Setimes.SR---A Reference Training Corpus of Serbian}.
\newblock In \emph{Proceedings of the Conference on Language Technologies \& Digital Humanities 2018 (JT-DH 2018)}, pages 11--17.

\bibitem[{Batsuren et~al.(2021)Batsuren, Bella, and Giunchiglia}]{batsuren-etal-2021-morphynet}
Khuyagbaatar Batsuren, G{\'a}bor Bella, and Fausto Giunchiglia. 2021.
\newblock \href {https://doi.org/10.18653/v1/2021.sigmorphon-1.5} {{M}orphy{N}et: a large multilingual database of derivational and inflectional morphology}.
\newblock In \emph{Proceedings of the 18th SIGMORPHON Workshop on Computational Research in Phonetics, Phonology, and Morphology}, pages 39--48, Online. Association for Computational Linguistics.

\bibitem[{Batsuren et~al.(2022)Batsuren, Goldman, Khalifa, Habash, Kiera{\'s}, Bella, Leonard, Nicolai, Gorman, Ate, Ryskina, Mielke, Budianskaya, El-Khaissi, Pimentel, Gasser, Lane, Raj, Coler, Samame, Camaiteri, Rojas, L{\'o}pez~Francis, Oncevay, L{\'o}pez~Bautista, Villegas, Hennigen, Ek, Guriel, Dirix, Bernardy, Scherbakov, Bayyr-ool, Anastasopoulos, Zariquiey, Sheifer, Ganieva, Cruz, Karah{\'o}{\v{g}}a, Markantonatou, Pavlidis, Plugaryov, Klyachko, Salehi, Angulo, Baxi, Krizhanovsky, Krizhanovskaya, Salesky, Vania, Ivanova, White, Maudslay, Valvoda, Zmigrod, Czarnowska, Nikkarinen, Salchak, Bhatt, Straughn, Liu, Washington, Pinter, Ataman, Wolinski, Suhardijanto, Yablonskaya, Stoehr, Dolatian, Nuriah, Ratan, Tyers, Ponti, Aiton, Arora, Hatcher, Kumar, Young, Rodionova, Yemelina, Andrushko, Marchenko, Mashkovtseva, Serova, Prud{'}hommeaux, Nepomniashchaya, Giunchiglia, Chodroff, Hulden, Silfverberg, McCarthy, Yarowsky, Cotterell, Tsarfaty, and Vylomova}]{batsuren-etal-2022-unimorph}
Khuyagbaatar Batsuren, Omer Goldman, Salam Khalifa, Nizar Habash, Witold Kiera{\'s}, G{\'a}bor Bella, Brian Leonard, Garrett Nicolai, Kyle Gorman, Yustinus~Ghanggo Ate, Maria Ryskina, Sabrina Mielke, Elena Budianskaya, Charbel El-Khaissi, Tiago Pimentel, Michael Gasser, William~Abbott Lane, Mohit Raj, Matt Coler, Jaime Rafael~Montoya Samame, Delio~Siticonatzi Camaiteri, Esa{\'u}~Zumaeta Rojas, Didier L{\'o}pez~Francis, Arturo Oncevay, Juan L{\'o}pez~Bautista, Gema Celeste~Silva Villegas, Lucas~Torroba Hennigen, Adam Ek, David Guriel, Peter Dirix, Jean-Philippe Bernardy, Andrey Scherbakov, Aziyana Bayyr-ool, Antonios Anastasopoulos, Roberto Zariquiey, Karina Sheifer, Sofya Ganieva, Hilaria Cruz, Ritv{\'a}n Karah{\'o}{\v{g}}a, Stella Markantonatou, George Pavlidis, Matvey Plugaryov, Elena Klyachko, Ali Salehi, Candy Angulo, Jatayu Baxi, Andrew Krizhanovsky, Natalia Krizhanovskaya, Elizabeth Salesky, Clara Vania, Sardana Ivanova, Jennifer White, Rowan~Hall Maudslay, Josef Valvoda, Ran Zmigrod, Paula Czarnowska,
  Irene Nikkarinen, Aelita Salchak, Brijesh Bhatt, Christopher Straughn, Zoey Liu, Jonathan~North Washington, Yuval Pinter, Duygu Ataman, Marcin Wolinski, Totok Suhardijanto, Anna Yablonskaya, Niklas Stoehr, Hossep Dolatian, Zahroh Nuriah, Shyam Ratan, Francis~M. Tyers, Edoardo~M. Ponti, Grant Aiton, Aryaman Arora, Richard~J. Hatcher, Ritesh Kumar, Jeremiah Young, Daria Rodionova, Anastasia Yemelina, Taras Andrushko, Igor Marchenko, Polina Mashkovtseva, Alexandra Serova, Emily Prud{'}hommeaux, Maria Nepomniashchaya, Fausto Giunchiglia, Eleanor Chodroff, Mans Hulden, Miikka Silfverberg, Arya~D. McCarthy, David Yarowsky, Ryan Cotterell, Reut Tsarfaty, and Ekaterina Vylomova. 2022.
\newblock \href {https://aclanthology.org/2022.lrec-1.89} {{U}ni{M}orph 4.0: {U}niversal {M}orphology}.
\newblock In \emph{Proceedings of the Thirteenth Language Resources and Evaluation Conference}, pages 840--855, Marseille, France. European Language Resources Association.

\bibitem[{Chen et~al.(2023{\natexlab{a}})Chen, Jiang, Ritter, and Xu}]{chen-etal-2023-frustratingly}
Yang Chen, Chao Jiang, Alan Ritter, and Wei Xu. 2023{\natexlab{a}}.
\newblock \href {https://doi.org/10.18653/v1/2023.findings-acl.357} {Frustratingly easy label projection for cross-lingual transfer}.
\newblock In \emph{Findings of the Association for Computational Linguistics: ACL 2023}, pages 5775--5796, Toronto, Canada. Association for Computational Linguistics.

\bibitem[{Chen et~al.(2023{\natexlab{b}})Chen, Shah, and Ritter}]{chen2023better}
Yang Chen, Vedaant Shah, and Alan Ritter. 2023{\natexlab{b}}.
\newblock \href {https://arxiv.org/pdf/2305.13582} {{Better Low-Resource Entity Recognition Through Translation and Annotation Fusion}}.
\newblock \emph{arXiv preprint arXiv:2305.13582}.

\bibitem[{Conneau et~al.(2020)Conneau, Khandelwal, Goyal, Chaudhary, Wenzek, Guzm{\'a}n, Grave, Ott, Zettlemoyer, and Stoyanov}]{conneau-etal-2020-unsupervised}
Alexis Conneau, Kartikay Khandelwal, Naman Goyal, Vishrav Chaudhary, Guillaume Wenzek, Francisco Guzm{\'a}n, Edouard Grave, Myle Ott, Luke Zettlemoyer, and Veselin Stoyanov. 2020.
\newblock \href {https://doi.org/10.18653/v1/2020.acl-main.747} {Unsupervised cross-lingual representation learning at scale}.
\newblock In \emph{Proceedings of the 58th Annual Meeting of the Association for Computational Linguistics}, pages 8440--8451, Online. Association for Computational Linguistics.

\bibitem[{Cucerzan and Yarowsky(1999)}]{cucerzan-yarowsky-1999-language}
Silviu Cucerzan and David Yarowsky. 1999.
\newblock \href {https://aclanthology.org/W99-0612} {Language independent named entity recognition combining morphological and contextual evidence}.
\newblock In \emph{1999 Joint {SIGDAT} Conference on Empirical Methods in Natural Language Processing and Very Large Corpora}.

\bibitem[{de~Marneffe et~al.(2021)de~Marneffe, Manning, Nivre, and Zeman}]{de-marneffe-etal-2021-universal}
Marie-Catherine de~Marneffe, Christopher~D. Manning, Joakim Nivre, and Daniel Zeman. 2021.
\newblock \href {https://doi.org/10.1162/coli_a_00402} {{U}niversal {D}ependencies}.
\newblock \emph{Computational Linguistics}, 47(2):255--308.

\bibitem[{Dione et~al.(2023{\natexlab{a}})Dione, Adelani, Nabende, Alabi, Sindane, Buzaaba, Muhammad, Emezue, Ogayo, Aremu, Gitau, Mbaye, Mukiibi, Sibanda, Dossou, Bukula, Mabuya, Tapo, Munkoh-Buabeng, victoire Memdjokam~Koagne, Kabore, Taylor, Kalipe, Macucwa, Marivate, Gwadabe, Elvis, Onyenwe, Atindogbe, Adelani, Akinade, Samuel, Nahimana, Musabeyezu, Niyomutabazi, Chimhenga, Gotosa, Mizha, Agbolo, Traore, Uchechukwu, Yusuf, Abdullahi, and Klakow}]{dione2023masakhapos}
Cheikh M.~Bamba Dione, David Adelani, Peter Nabende, Jesujoba Alabi, Thapelo Sindane, Happy Buzaaba, Shamsuddeen~Hassan Muhammad, Chris~Chinenye Emezue, Perez Ogayo, Anuoluwapo Aremu, Catherine Gitau, Derguene Mbaye, Jonathan Mukiibi, Blessing Sibanda, Bonaventure F.~P. Dossou, Andiswa Bukula, Rooweither Mabuya, Allahsera~Auguste Tapo, Edwin Munkoh-Buabeng, victoire Memdjokam~Koagne, Fatoumata~Ouoba Kabore, Amelia Taylor, Godson Kalipe, Tebogo Macucwa, Vukosi Marivate, Tajuddeen Gwadabe, Mboning~Tchiaze Elvis, Ikechukwu Onyenwe, Gratien Atindogbe, Tolulope Adelani, Idris Akinade, Olanrewaju Samuel, Marien Nahimana, Théogène Musabeyezu, Emile Niyomutabazi, Ester Chimhenga, Kudzai Gotosa, Patrick Mizha, Apelete Agbolo, Seydou Traore, Chinedu Uchechukwu, Aliyu Yusuf, Muhammad Abdullahi, and Dietrich Klakow. 2023{\natexlab{a}}.
\newblock \href {http://arxiv.org/abs/2305.13989} {Masakhapos: Part-of-speech tagging for typologically diverse african languages}.

\bibitem[{Dione et~al.(2023{\natexlab{b}})Dione, Adelani, Nabende, Alabi, Sindane, Buzaaba, Muhammad, Emezue, Ogayo, Aremu, Gitau, Mbaye, Mukiibi, Sibanda, Dossou, Bukula, Mabuya, Tapo, Munkoh-Buabeng, Memdjokam~Koagne, Ouoba~Kabore, Taylor, Kalipe, Macucwa, Marivate, Gwadabe, Elvis, Onyenwe, Atindogbe, Adelani, Akinade, Samuel, Nahimana, Musabeyezu, Niyomutabazi, Chimhenga, Gotosa, Mizha, Agbolo, Traore, Uchechukwu, Yusuf, Abdullahi, and Klakow}]{dione-etal-2023-masakhapos}
Cheikh M.~Bamba Dione, David~Ifeoluwa Adelani, Peter Nabende, Jesujoba Alabi, Thapelo Sindane, Happy Buzaaba, Shamsuddeen~Hassan Muhammad, Chris~Chinenye Emezue, Perez Ogayo, Anuoluwapo Aremu, Catherine Gitau, Derguene Mbaye, Jonathan Mukiibi, Blessing Sibanda, Bonaventure F.~P. Dossou, Andiswa Bukula, Rooweither Mabuya, Allahsera~Auguste Tapo, Edwin Munkoh-Buabeng, Victoire Memdjokam~Koagne, Fatoumata Ouoba~Kabore, Amelia Taylor, Godson Kalipe, Tebogo Macucwa, Vukosi Marivate, Tajuddeen Gwadabe, Mboning~Tchiaze Elvis, Ikechukwu Onyenwe, Gratien Atindogbe, Tolulope Adelani, Idris Akinade, Olanrewaju Samuel, Marien Nahimana, Th{\'e}og{\`e}ne Musabeyezu, Emile Niyomutabazi, Ester Chimhenga, Kudzai Gotosa, Patrick Mizha, Apelete Agbolo, Seydou Traore, Chinedu Uchechukwu, Aliyu Yusuf, Muhammad Abdullahi, and Dietrich Klakow. 2023{\natexlab{b}}.
\newblock \href {https://doi.org/10.18653/v1/2023.acl-long.609} {{M}asakha{POS}: Part-of-speech tagging for typologically diverse {A}frican languages}.
\newblock In \emph{Proceedings of the 61st Annual Meeting of the Association for Computational Linguistics (Volume 1: Long Papers)}, pages 10883--10900, Toronto, Canada. Association for Computational Linguistics.

\bibitem[{Grishman(2019)}]{grishman_2019}
Ralph Grishman. 2019.
\newblock \href {https://doi.org/10.1017/S1351324919000512} {Twenty-five years of information extraction}.
\newblock \emph{Natural Language Engineering}, 25(6):677–692.

\bibitem[{Hvingelby et~al.(2020)Hvingelby, Pauli, Barrett, Rosted, Lidegaard, and S{\o}gaard}]{hvingelby-etal-2020-dane}
Rasmus Hvingelby, Amalie~Brogaard Pauli, Maria Barrett, Christina Rosted, Lasse~Malm Lidegaard, and Anders S{\o}gaard. 2020.
\newblock \href {https://aclanthology.org/2020.lrec-1.565} {{D}a{NE}: A named entity resource for {D}anish}.
\newblock In \emph{Proceedings of the Twelfth Language Resources and Evaluation Conference}, pages 4597--4604, Marseille, France. European Language Resources Association.

\bibitem[{{International Organization for Standardization}(1998)}]{ISO639-2}
{International Organization for Standardization}. 1998.
\newblock {Codes for the representation of names of languages---Part 2: alpha-3 code}.
\newblock Standard, International Organization for Standardization, Geneva, CH.

\bibitem[{{International Organization for Standardization}(2002)}]{ISO639-1}
{International Organization for Standardization}. 2002.
\newblock {Codes for the representation of names of languages---Part 1: Alpha-2 code}.
\newblock Standard, International Organization for Standardization, Geneva, CH.

\bibitem[{Johannsen et~al.(2015)Johannsen, Alonso, and Plank}]{johannsen-etal-2015-universal}
Anders Johannsen, H{\'e}ctor~Mart{\'\i}nez Alonso, and Barbara Plank. 2015.
\newblock \href {http://tlt14.ipipan.waw.pl/files/4914/4974/3227/TLT14_proceedings.pdf#page=164} {{Universal Dependencies for Danish}}.
\newblock In \emph{International Workshop on Treebanks and Linguistic Theories (TLT14)}, page 157.

\bibitem[{J{\o}rgensen et~al.(2020)J{\o}rgensen, Aasmoe, Ruud~Husev{\aa}g, {\O}vrelid, and Velldal}]{jorgensen-etal-2020-norne}
Fredrik J{\o}rgensen, Tobias Aasmoe, Anne-Stine Ruud~Husev{\aa}g, Lilja {\O}vrelid, and Erik Velldal. 2020.
\newblock \href {https://aclanthology.org/2020.lrec-1.559} {{N}or{NE}: Annotating named entities for {N}orwegian}.
\newblock In \emph{Proceedings of the Twelfth Language Resources and Evaluation Conference}, pages 4547--4556, Marseille, France. European Language Resources Association.

\bibitem[{Kahane et~al.(2021)Kahane, Caron, Strickland, and Gerdes}]{kahane-etal-2021-annotation}
Sylvain Kahane, Bernard Caron, Emmett Strickland, and Kim Gerdes. 2021.
\newblock \href {https://aclanthology.org/2021.tlt-1.4} {Annotation guidelines of {UD} and {SUD} treebanks for spoken corpora: A proposal}.
\newblock In \emph{Proceedings of the 20th International Workshop on Treebanks and Linguistic Theories (TLT, SyntaxFest 2021)}, pages 35--47, Sofia, Bulgaria. Association for Computational Linguistics.

\bibitem[{Khalid et~al.(2008)Khalid, Jijkoun, and De~Rijke}]{khalid2008impact}
Mahboob~Alam Khalid, Valentin Jijkoun, and Maarten De~Rijke. 2008.
\newblock The impact of named entity normalization on information retrieval for question answering.
\newblock In \emph{European Conference on Information Retrieval}, pages 705--710. Springer.

\bibitem[{Kirov et~al.(2018)Kirov, Cotterell, Sylak-Glassman, Walther, Vylomova, Xia, Faruqui, Mielke, McCarthy, K{\"u}bler, Yarowsky, Eisner, and Hulden}]{kirov-etal-2018-unimorph}
Christo Kirov, Ryan Cotterell, John Sylak-Glassman, G{\'e}raldine Walther, Ekaterina Vylomova, Patrick Xia, Manaal Faruqui, Sabrina~J. Mielke, Arya McCarthy, Sandra K{\"u}bler, David Yarowsky, Jason Eisner, and Mans Hulden. 2018.
\newblock \href {https://aclanthology.org/L18-1293} {{U}ni{M}orph 2.0: {U}niversal {M}orphology}.
\newblock In \emph{Proceedings of the Eleventh International Conference on Language Resources and Evaluation ({LREC} 2018)}, Miyazaki, Japan. European Language Resources Association (ELRA).

\bibitem[{Lample et~al.(2016)Lample, Ballesteros, Subramanian, Kawakami, and Dyer}]{lample-etal-2016-neural}
Guillaume Lample, Miguel Ballesteros, Sandeep Subramanian, Kazuya Kawakami, and Chris Dyer. 2016.
\newblock \href {https://doi.org/10.18653/v1/N16-1030} {Neural architectures for named entity recognition}.
\newblock In \emph{Proceedings of the 2016 Conference of the North {A}merican Chapter of the Association for Computational Linguistics: Human Language Technologies}, pages 260--270, San Diego, California. Association for Computational Linguistics.

\bibitem[{Lignos et~al.(2022)Lignos, Holley, Palen-Michel, and S{\"a}lev{\"a}}]{lignos-etal-2022-toward}
Constantine Lignos, Nolan Holley, Chester Palen-Michel, and Jonne S{\"a}lev{\"a}. 2022.
\newblock \href {https://doi.org/10.18653/v1/2022.findings-acl.44} {Toward more meaningful resources for lower-resourced languages}.
\newblock In \emph{Findings of the Association for Computational Linguistics: ACL 2022}, pages 523--532, Dublin, Ireland. Association for Computational Linguistics.

\bibitem[{Ljube{\v{s}}i{\'c} et~al.(2016)Ljube{\v{s}}i{\'c}, Klubi{\v{c}}ka, Agi{\'c}, and Jazbec}]{ljubesic-etal-2016-new}
Nikola Ljube{\v{s}}i{\'c}, Filip Klubi{\v{c}}ka, {\v{Z}}eljko Agi{\'c}, and Ivo-Pavao Jazbec. 2016.
\newblock \href {https://aclanthology.org/L16-1676} {New inflectional lexicons and training corpora for improved morphosyntactic annotation of {C}roatian and {S}erbian}.
\newblock In \emph{Proceedings of the Tenth International Conference on Language Resources and Evaluation ({LREC}'16)}, pages 4264--4270, Portoro{\v{z}}, Slovenia. European Language Resources Association (ELRA).

\bibitem[{Luoma et~al.(2020)Luoma, Oinonen, Pyyk{\"o}nen, Laippala, and Pyysalo}]{luoma-etal-2020-broad}
Jouni Luoma, Miika Oinonen, Maria Pyyk{\"o}nen, Veronika Laippala, and Sampo Pyysalo. 2020.
\newblock \href {https://aclanthology.org/2020.lrec-1.567} {A broad-coverage corpus for {F}innish named entity recognition}.
\newblock In \emph{Proceedings of the Twelfth Language Resources and Evaluation Conference}, pages 4615--4624, Marseille, France. European Language Resources Association.

\bibitem[{Ma et~al.(2023)Ma, Wu, Jiang, Karlsson, Zhao, and Lin}]{ma2023colada}
Tingting Ma, Qianhui Wu, Huiqiang Jiang, B{\"o}rje~F. Karlsson, Tiejun Zhao, and Chin-Yew Lin. 2023.
\newblock \href {https://doi.org/10.18653/v1/2023.acl-long.330} {{C}o{L}a{D}a: A collaborative label denoising framework for cross-lingual named entity recognition}.
\newblock In \emph{Proceedings of the 61st Annual Meeting of the Association for Computational Linguistics (Volume 1: Long Papers)}. Association for Computational Linguistics.

\bibitem[{Malmasi et~al.(2022{\natexlab{a}})Malmasi, Fang, Fetahu, Kar, and Rokhlenko}]{malmasi-etal-2022-multiconer}
Shervin Malmasi, Anjie Fang, Besnik Fetahu, Sudipta Kar, and Oleg Rokhlenko. 2022{\natexlab{a}}.
\newblock \href {https://aclanthology.org/2022.coling-1.334} {{M}ulti{C}o{NER}: A large-scale multilingual dataset for complex named entity recognition}.
\newblock In \emph{Proceedings of the 29th International Conference on Computational Linguistics}, pages 3798--3809, Gyeongju, Republic of Korea. International Committee on Computational Linguistics.

\bibitem[{Malmasi et~al.(2022{\natexlab{b}})Malmasi, Fang, Fetahu, Kar, and Rokhlenko}]{malmasi-etal-2022-semeval}
Shervin Malmasi, Anjie Fang, Besnik Fetahu, Sudipta Kar, and Oleg Rokhlenko. 2022{\natexlab{b}}.
\newblock \href {https://doi.org/10.18653/v1/2022.semeval-1.196} {{S}em{E}val-2022 task 11: Multilingual complex named entity recognition ({M}ulti{C}o{NER})}.
\newblock In \emph{Proceedings of the 16th International Workshop on Semantic Evaluation (SemEval-2022)}, pages 1412--1437, Seattle, United States. Association for Computational Linguistics.

\bibitem[{Mayhew and Roth(2018)}]{mayhew-roth-2018-talen}
Stephen Mayhew and Dan Roth. 2018.
\newblock \href {https://doi.org/10.18653/v1/P18-4014} {{TALEN}: Tool for annotation of low-resource {EN}tities}.
\newblock In \emph{Proceedings of {ACL} 2018, System Demonstrations}, pages 80--86, Melbourne, Australia. Association for Computational Linguistics.

\bibitem[{McCarthy et~al.(2020)McCarthy, Kirov, Grella, Nidhi, Xia, Gorman, Vylomova, Mielke, Nicolai, Silfverberg, Arkhangelskiy, Krizhanovsky, Krizhanovsky, Klyachko, Sorokin, Mansfield, Ern{\v{s}}treits, Pinter, Jacobs, Cotterell, Hulden, and Yarowsky}]{mccarthy-etal-2020-unimorph}
Arya~D. McCarthy, Christo Kirov, Matteo Grella, Amrit Nidhi, Patrick Xia, Kyle Gorman, Ekaterina Vylomova, Sabrina~J. Mielke, Garrett Nicolai, Miikka Silfverberg, Timofey Arkhangelskiy, Nataly Krizhanovsky, Andrew Krizhanovsky, Elena Klyachko, Alexey Sorokin, John Mansfield, Valts Ern{\v{s}}treits, Yuval Pinter, Cassandra~L. Jacobs, Ryan Cotterell, Mans Hulden, and David Yarowsky. 2020.
\newblock \href {https://aclanthology.org/2020.lrec-1.483} {{U}ni{M}orph 3.0: {U}niversal {M}orphology}.
\newblock In \emph{Proceedings of the Twelfth Language Resources and Evaluation Conference}, pages 3922--3931, Marseille, France. European Language Resources Association.

\bibitem[{McDonald et~al.(2013)McDonald, Nivre, Quirmbach-Brundage, Goldberg, Das, Ganchev, Hall, Petrov, Zhang, T{\"a}ckstr{\"o}m, Bedini, Bertomeu~Castell{\'o}, and Lee}]{mcdonald-etal-2013-universal}
Ryan McDonald, Joakim Nivre, Yvonne Quirmbach-Brundage, Yoav Goldberg, Dipanjan Das, Kuzman Ganchev, Keith Hall, Slav Petrov, Hao Zhang, Oscar T{\"a}ckstr{\"o}m, Claudia Bedini, N{\'u}ria Bertomeu~Castell{\'o}, and Jungmee Lee. 2013.
\newblock \href {https://aclanthology.org/P13-2017} {{U}niversal {D}ependency annotation for multilingual parsing}.
\newblock In \emph{Proceedings of the 51st Annual Meeting of the Association for Computational Linguistics (Volume 2: Short Papers)}, pages 92--97, Sofia, Bulgaria. Association for Computational Linguistics.

\bibitem[{Moll{\'a} et~al.(2006)Moll{\'a}, van Zaanen, and Smith}]{molla-etal-2006-named}
Diego Moll{\'a}, Menno van Zaanen, and Daniel Smith. 2006.
\newblock \href {https://aclanthology.org/U06-1009} {Named entity recognition for question answering}.
\newblock In \emph{Proceedings of the Australasian Language Technology Workshop 2006}, pages 51--58, Sydney, Australia.

\bibitem[{Muennighoff et~al.(2023)Muennighoff, Wang, Sutawika, Roberts, Biderman, Le~Scao, Bari, Shen, Yong, Schoelkopf, Tang, Radev, Aji, Almubarak, Albanie, Alyafeai, Webson, Raff, and Raffel}]{muennighoff-etal-2023-crosslingual}
Niklas Muennighoff, Thomas Wang, Lintang Sutawika, Adam Roberts, Stella Biderman, Teven Le~Scao, M~Saiful Bari, Sheng Shen, Zheng~Xin Yong, Hailey Schoelkopf, Xiangru Tang, Dragomir Radev, Alham~Fikri Aji, Khalid Almubarak, Samuel Albanie, Zaid Alyafeai, Albert Webson, Edward Raff, and Colin Raffel. 2023.
\newblock \href {https://doi.org/10.18653/v1/2023.acl-long.891} {Crosslingual generalization through multitask finetuning}.
\newblock In \emph{Proceedings of the 61st Annual Meeting of the Association for Computational Linguistics (Volume 1: Long Papers)}, pages 15991--16111, Toronto, Canada. Association for Computational Linguistics.

\bibitem[{Muischnek and M{\"u}{\"u}risep(2023)}]{muischnek-muurisep-2023-named}
Kadri Muischnek and Kaili M{\"u}{\"u}risep. 2023.
\newblock \href {https://aclanthology.org/2023.nodalida-1.19} {Named entity layer in {E}stonian {UD} treebanks}.
\newblock In \emph{Proceedings of the 24th Nordic Conference on Computational Linguistics (NoDaLiDa)}, pages 179--184, T{\'o}rshavn, Faroe Islands. University of Tartu Library.

\bibitem[{Nivre et~al.(2016)Nivre, de~Marneffe, Ginter, Goldberg, Haji{\v{c}}, Manning, McDonald, Petrov, Pyysalo, Silveira, Tsarfaty, and Zeman}]{nivre-etal-2016-universal}
Joakim Nivre, Marie-Catherine de~Marneffe, Filip Ginter, Yoav Goldberg, Jan Haji{\v{c}}, Christopher~D. Manning, Ryan McDonald, Slav Petrov, Sampo Pyysalo, Natalia Silveira, Reut Tsarfaty, and Daniel Zeman. 2016.
\newblock \href {https://aclanthology.org/L16-1262} {{U}niversal {D}ependencies v1: A multilingual treebank collection}.
\newblock In \emph{Proceedings of the Tenth International Conference on Language Resources and Evaluation ({LREC}'16)}, pages 1659--1666, Portoro{\v{z}}, Slovenia. European Language Resources Association (ELRA).

\bibitem[{Nivre et~al.(2020)Nivre, de~Marneffe, Ginter, Haji{\v{c}}, Manning, Pyysalo, Schuster, Tyers, and Zeman}]{nivre-etal-2020-universal}
Joakim Nivre, Marie-Catherine de~Marneffe, Filip Ginter, Jan Haji{\v{c}}, Christopher~D. Manning, Sampo Pyysalo, Sebastian Schuster, Francis Tyers, and Daniel Zeman. 2020.
\newblock \href {https://aclanthology.org/2020.lrec-1.497} {{U}niversal {D}ependencies v2: An evergrowing multilingual treebank collection}.
\newblock In \emph{Proceedings of the Twelfth Language Resources and Evaluation Conference}, pages 4034--4043, Marseille, France. European Language Resources Association.

\bibitem[{Nothman et~al.(2013)Nothman, Ringland, Radford, Murphy, and Curran}]{nothman2013learning}
Joel Nothman, Nicky Ringland, Will Radford, Tara Murphy, and James~R. Curran. 2013.
\newblock \href {https://www.sciencedirect.com/science/article/pii/S0004370212000276/pdf?md5=e8d24dc17740f10183b96672d507452c&pid=1-s2.0-S0004370212000276-main.pdf&_valck=1} {{Learning multilingual named entity recognition from Wikipedia}}.
\newblock \emph{Artificial Intelligence}, 194:151--175.

\bibitem[{Pan et~al.(2017)Pan, Zhang, May, Nothman, Knight, and Ji}]{pan-etal-2017-cross}
Xiaoman Pan, Boliang Zhang, Jonathan May, Joel Nothman, Kevin Knight, and Heng Ji. 2017.
\newblock \href {https://doi.org/10.18653/v1/P17-1178} {Cross-lingual name tagging and linking for 282 languages}.
\newblock In \emph{Proceedings of the 55th Annual Meeting of the Association for Computational Linguistics (Volume 1: Long Papers)}, pages 1946--1958, Vancouver, Canada. Association for Computational Linguistics.

\bibitem[{Petrov et~al.(2012)Petrov, Das, and McDonald}]{petrov-etal-2012-universal}
Slav Petrov, Dipanjan Das, and Ryan McDonald. 2012.
\newblock \href {http://www.lrec-conf.org/proceedings/lrec2012/pdf/274_Paper.pdf} {A universal part-of-speech tagset}.
\newblock In \emph{Proceedings of the Eighth International Conference on Language Resources and Evaluation ({LREC}'12)}, pages 2089--2096, Istanbul, Turkey. European Language Resources Association (ELRA).

\bibitem[{Plank(2019)}]{plank-2019-neural}
Barbara Plank. 2019.
\newblock \href {https://aclanthology.org/W19-6143} {Neural cross-lingual transfer and limited annotated data for named entity recognition in {D}anish}.
\newblock In \emph{Proceedings of the 22nd Nordic Conference on Computational Linguistics}, pages 370--375, Turku, Finland. Link{\"o}ping University Electronic Press.

\bibitem[{Plank(2021)}]{plank-2021-cross}
Barbara Plank. 2021.
\newblock \href {https://doi.org/10.18653/v1/2021.findings-acl.158} {Cross-lingual cross-domain nested named entity evaluation on {E}nglish web texts}.
\newblock In \emph{Findings of the Association for Computational Linguistics: ACL-IJCNLP 2021}, pages 1808--1815, Online. Association for Computational Linguistics.

\bibitem[{Plank et~al.(2020)Plank, Jensen, and van~der Goot}]{plank-etal-2020-DaN}
Barbara Plank, Kristian~N{\o}rgaard Jensen, and Rob van~der Goot. 2020.
\newblock \href {https://doi.org/10.18653/v1/2020.coling-main.583} {{D}a{N}+: {D}anish nested named entities and lexical normalization}.
\newblock In \emph{Proceedings of the 28th International Conference on Computational Linguistics}, pages 6649--6662, Barcelona, Spain (Online). International Committee on Computational Linguistics.

\bibitem[{Qi and Yasuoka(2019)}]{ud_chinese_gsdsimp}
Peng Qi and Koichi Yasuoka. 2019.
\newblock {UD\_Chinese-GSDSimp}.
\newblock \url{https://github.com/UniversalDependencies/UD_Chinese-GSDSimp}.

\bibitem[{Rademaker et~al.(2017)Rademaker, Chalub, Real, Freitas, Bick, and de~Paiva}]{rademaker-etal-2017-universal}
Alexandre Rademaker, Fabricio Chalub, Livy Real, Cl{\'a}udia Freitas, Eckhard Bick, and Valeria de~Paiva. 2017.
\newblock \href {https://aclanthology.org/W17-6523} {{U}niversal {D}ependencies for {P}ortuguese}.
\newblock In \emph{Proceedings of the Fourth International Conference on Dependency Linguistics (Depling 2017)}, pages 197--206, Pisa, Italy. Link{\"o}ping University Electronic Press.

\bibitem[{Riabi et~al.(2023)Riabi, Mahamdi, and Seddah}]{riabi-etal-2023-enriching}
Arij Riabi, Menel Mahamdi, and Djam{\'e} Seddah. 2023.
\newblock \href {https://doi.org/10.18653/v1/2023.law-1.26} {Enriching the {NA}rabizi treebank: A multifaceted approach to supporting an under-resourced language}.
\newblock In \emph{Proceedings of the 17th Linguistic Annotation Workshop (LAW-XVII)}, pages 266--278, Toronto, Canada. Association for Computational Linguistics.

\bibitem[{Ruder et~al.(2019)Ruder, S{\o}gaard, and Vuli{\'c}}]{ruder-etal-2019-unsupervised}
Sebastian Ruder, Anders S{\o}gaard, and Ivan Vuli{\'c}. 2019.
\newblock \href {https://doi.org/10.18653/v1/P19-4007} {Unsupervised cross-lingual representation learning}.
\newblock In \emph{Proceedings of the 57th Annual Meeting of the Association for Computational Linguistics: Tutorial Abstracts}, pages 31--38, Florence, Italy. Association for Computational Linguistics.

\bibitem[{Samard{\v{z}}i{\'c} et~al.(2017)Samard{\v{z}}i{\'c}, Starovi{\'c}, Agi{\'c}, and Ljube{\v{s}}i{\'c}}]{samardzic-etal-2017-universal}
Tanja Samard{\v{z}}i{\'c}, Mirjana Starovi{\'c}, {\v{Z}}eljko Agi{\'c}, and Nikola Ljube{\v{s}}i{\'c}. 2017.
\newblock \href {https://doi.org/10.18653/v1/W17-1407} {{U}niversal {D}ependencies for {S}erbian in comparison with {C}roatian and other {S}lavic languages}.
\newblock In \emph{Proceedings of the 6th Workshop on {B}alto-{S}lavic Natural Language Processing}, pages 39--44, Valencia, Spain. Association for Computational Linguistics.

\bibitem[{Samson and C{\"o}ltekin(2020)}]{ud_tagalog_trg}
Stephanie Samson and Cagr{\i} C{\"o}ltekin. 2020.
\newblock {UD\_Tagalog-TRG}.
\newblock \url{https://github.com/UniversalDependencies/UD_Tagalog-TRG}.

\bibitem[{Savary et~al.(2023)Savary, Ben~Khelil, Ramisch, Giouli, Barbu~Mititelu, Hadj~Mohamed, Krstev, Liebeskind, Xu, Stymne, G{\"u}ng{\"o}r, Pickard, Guillaume, Bej{\v{c}}ek, Bhatia, Candito, Gantar, I{\~n}urrieta, Gatt, Kovalevskaite, Lichte, Ljube{\v{s}}i{\'c}, Monti, Parra~Escart{\'\i}n, Shamsfard, Stoyanova, Vincze, and Walsh}]{savary-etal-2023-parseme}
Agata Savary, Cherifa Ben~Khelil, Carlos Ramisch, Voula Giouli, Verginica Barbu~Mititelu, Najet Hadj~Mohamed, Cvetana Krstev, Chaya Liebeskind, Hongzhi Xu, Sara Stymne, Tunga G{\"u}ng{\"o}r, Thomas Pickard, Bruno Guillaume, Eduard Bej{\v{c}}ek, Archna Bhatia, Marie Candito, Polona Gantar, Uxoa I{\~n}urrieta, Albert Gatt, Jolanta Kovalevskaite, Timm Lichte, Nikola Ljube{\v{s}}i{\'c}, Johanna Monti, Carla Parra~Escart{\'\i}n, Mehrnoush Shamsfard, Ivelina Stoyanova, Veronika Vincze, and Abigail Walsh. 2023.
\newblock \href {https://doi.org/10.18653/v1/2023.mwe-1.6} {{PARSEME} corpus release 1.3}.
\newblock In \emph{Proceedings of the 19th Workshop on Multiword Expressions (MWE 2023)}, pages 24--35, Dubrovnik, Croatia. Association for Computational Linguistics.

\bibitem[{Seddah et~al.(2020)Seddah, Essaidi, Fethi, Futeral, Muller, Ortiz~Su{\'a}rez, Sagot, and Srivastava}]{seddah-etal-2020-building}
Djam{\'e} Seddah, Farah Essaidi, Amal Fethi, Matthieu Futeral, Benjamin Muller, Pedro~Javier Ortiz~Su{\'a}rez, Beno{\^\i}t Sagot, and Abhishek Srivastava. 2020.
\newblock \href {https://doi.org/10.18653/v1/2020.acl-main.107} {Building a user-generated content {N}orth-{A}frican {A}rabizi treebank: Tackling hell}.
\newblock In \emph{Proceedings of the 58th Annual Meeting of the Association for Computational Linguistics}, pages 1139--1150, Online. Association for Computational Linguistics.

\bibitem[{Shen et~al.(2016)Shen, McDonald, Zeman, and Qi}]{ud_chinese_gsd}
Mo~Shen, Ryan McDonald, Daniel Zeman, and Peng Qi. 2016.
\newblock {UD\_Chinese-GSD}.
\newblock \url{https://github.com/UniversalDependencies/UD_Chinese-GSD}.

\bibitem[{Silveira et~al.(2014)Silveira, Dozat, de~Marneffe, Bowman, Connor, Bauer, and Manning}]{silveira-etal-2014-gold}
Natalia Silveira, Timothy Dozat, Marie-Catherine de~Marneffe, Samuel Bowman, Miriam Connor, John Bauer, and Chris Manning. 2014.
\newblock \href {http://www.lrec-conf.org/proceedings/lrec2014/pdf/1089_Paper.pdf} {A gold standard dependency corpus for {E}nglish}.
\newblock In \emph{Proceedings of the Ninth International Conference on Language Resources and Evaluation ({LREC}'14)}, pages 2897--2904, Reykjavik, Iceland. European Language Resources Association (ELRA).

\bibitem[{Strassel and Tracey(2016)}]{strassel-tracey-2016-lorelei}
Stephanie Strassel and Jennifer Tracey. 2016.
\newblock \href {https://aclanthology.org/L16-1521} {{LORELEI} language packs: Data, tools, and resources for technology development in low resource languages}.
\newblock In \emph{Proceedings of the Tenth International Conference on Language Resources and Evaluation ({LREC}'16)}, pages 3273--3280, Portoro{\v{z}}, Slovenia. European Language Resources Association (ELRA).

\bibitem[{Tedeschi et~al.(2021)Tedeschi, Maiorca, Campolungo, Cecconi, and Navigli}]{tedeschi-etal-2021-wikineural-combined}
Simone Tedeschi, Valentino Maiorca, Niccol{\`o} Campolungo, Francesco Cecconi, and Roberto Navigli. 2021.
\newblock \href {https://doi.org/10.18653/v1/2021.findings-emnlp.215} {{W}iki{NE}u{R}al: {C}ombined neural and knowledge-based silver data creation for multilingual {NER}}.
\newblock In \emph{Findings of the Association for Computational Linguistics: EMNLP 2021}, pages 2521--2533, Punta Cana, Dominican Republic. Association for Computational Linguistics.

\bibitem[{Tjong Kim~Sang(2002)}]{tjong-kim-sang-2002-introduction}
Erik~F. Tjong Kim~Sang. 2002.
\newblock \href {https://aclanthology.org/W02-2024} {Introduction to the {C}o{NLL}-2002 shared task: Language-independent named entity recognition}.
\newblock In \emph{{COLING}-02: The 6th Conference on Natural Language Learning 2002 ({C}o{NLL}-2002)}.

\bibitem[{Tjong Kim~Sang and De~Meulder(2003)}]{tjong-kim-sang-de-meulder-2003-introduction}
Erik~F. Tjong Kim~Sang and Fien De~Meulder. 2003.
\newblock \href {https://aclanthology.org/W03-0419} {Introduction to the {C}o{NLL}-2003 shared task: Language-independent named entity recognition}.
\newblock In \emph{Proceedings of the Seventh Conference on Natural Language Learning at {HLT}-{NAACL} 2003}, pages 142--147.

\bibitem[{Wolf et~al.(2020)Wolf, Debut, Sanh, Chaumond, Delangue, Moi, Cistac, Rault, Louf, Funtowicz, Davison, Shleifer, von Platen, Ma, Jernite, Plu, Xu, Le~Scao, Gugger, Drame, Lhoest, and Rush}]{wolf-etal-2020-transformers}
Thomas Wolf, Lysandre Debut, Victor Sanh, Julien Chaumond, Clement Delangue, Anthony Moi, Pierric Cistac, Tim Rault, Remi Louf, Morgan Funtowicz, Joe Davison, Sam Shleifer, Patrick von Platen, Clara Ma, Yacine Jernite, Julien Plu, Canwen Xu, Teven Le~Scao, Sylvain Gugger, Mariama Drame, Quentin Lhoest, and Alexander Rush. 2020.
\newblock \href {https://doi.org/10.18653/v1/2020.emnlp-demos.6} {Transformers: State-of-the-art natural language processing}.
\newblock In \emph{Proceedings of the 2020 Conference on Empirical Methods in Natural Language Processing: System Demonstrations}, pages 38--45, Online. Association for Computational Linguistics.

\bibitem[{Wu et~al.(2020{\natexlab{a}})Wu, Lin, Karlsson, Lou, and Huang}]{wu-etal-2020-single}
Qianhui Wu, Zijia Lin, B{\"o}rje Karlsson, Jian-Guang Lou, and Biqing Huang. 2020{\natexlab{a}}.
\newblock \href {https://doi.org/10.18653/v1/2020.acl-main.581} {Single-/multi-source cross-lingual {NER} via teacher-student learning on unlabeled data in target language}.
\newblock In \emph{Proceedings of the 58th Annual Meeting of the Association for Computational Linguistics}, pages 6505--6514, Online. Association for Computational Linguistics.

\bibitem[{Wu et~al.(2020{\natexlab{b}})Wu, Lin, Karlsson, Huang, and Lou}]{wu-etal-2020-unitrans}
Qianhui Wu, Zijia Lin, Börje~F. Karlsson, Biqing Huang, and Jian-Guang Lou. 2020{\natexlab{b}}.
\newblock \href {https://doi.org/10.24963/ijcai.2020/543} {Unitrans : Unifying model transfer and data transfer for cross-lingual named entity recognition with unlabeled data}.
\newblock In \emph{Proceedings of the Twenty-Ninth International Joint Conference on Artificial Intelligence, {IJCAI-20}}. International Joint Conferences on Artificial Intelligence Organization.

\bibitem[{Wu and Dredze(2020)}]{wu-dredze-2020-languages}
Shijie Wu and Mark Dredze. 2020.
\newblock \href {https://doi.org/10.18653/v1/2020.repl4nlp-1.16} {Are all languages created equal in multilingual {BERT}?}
\newblock In \emph{Proceedings of the 5th Workshop on Representation Learning for NLP}, pages 120--130, Online. Association for Computational Linguistics.

\bibitem[{Zeman(2017)}]{zeman-2017-slovak}
Daniel Zeman. 2017.
\newblock \href {https://sciendo.com/pdf/10.1515/jazcas-2017-0048} {Slovak dependency treebank in universal dependencies}.
\newblock \emph{Journal of Linguistics/Jazykovedn{\`y} casopis}, 68(2):385--395.

\bibitem[{Zeman et~al.(2017)Zeman, Popel, Straka, Haji{\v{c}}, Nivre, Ginter, Luotolahti, Pyysalo, Petrov, Potthast, Tyers, Badmaeva, Gokirmak, Nedoluzhko, Cinkov{\'a}, Haji{\v{c}}~jr., Hlav{\'a}{\v{c}}ov{\'a}, Kettnerov{\'a}, Ure{\v{s}}ov{\'a}, Kanerva, Ojala, Missil{\"a}, Manning, Schuster, Reddy, Taji, Habash, Leung, de~Marneffe, Sanguinetti, Simi, Kanayama, de~Paiva, Droganova, Mart{\'\i}nez~Alonso, {\c{C}}{\"o}ltekin, Sulubacak, Uszkoreit, Macketanz, Burchardt, Harris, Marheinecke, Rehm, Kayadelen, Attia, Elkahky, Yu, Pitler, Lertpradit, Mandl, Kirchner, Alcalde, Strnadov{\'a}, Banerjee, Manurung, Stella, Shimada, Kwak, Mendon{\c{c}}a, Lando, Nitisaroj, and Li}]{zeman-etal-2017-conll}
Daniel Zeman, Martin Popel, Milan Straka, Jan Haji{\v{c}}, Joakim Nivre, Filip Ginter, Juhani Luotolahti, Sampo Pyysalo, Slav Petrov, Martin Potthast, Francis Tyers, Elena Badmaeva, Memduh Gokirmak, Anna Nedoluzhko, Silvie Cinkov{\'a}, Jan Haji{\v{c}}~jr., Jaroslava Hlav{\'a}{\v{c}}ov{\'a}, V{\'a}clava Kettnerov{\'a}, Zde{\v{n}}ka Ure{\v{s}}ov{\'a}, Jenna Kanerva, Stina Ojala, Anna Missil{\"a}, Christopher~D. Manning, Sebastian Schuster, Siva Reddy, Dima Taji, Nizar Habash, Herman Leung, Marie-Catherine de~Marneffe, Manuela Sanguinetti, Maria Simi, Hiroshi Kanayama, Valeria de~Paiva, Kira Droganova, H{\'e}ctor Mart{\'\i}nez~Alonso, {\c{C}}a{\u{g}}r{\i} {\c{C}}{\"o}ltekin, Umut Sulubacak, Hans Uszkoreit, Vivien Macketanz, Aljoscha Burchardt, Kim Harris, Katrin Marheinecke, Georg Rehm, Tolga Kayadelen, Mohammed Attia, Ali Elkahky, Zhuoran Yu, Emily Pitler, Saran Lertpradit, Michael Mandl, Jesse Kirchner, Hector~Fernandez Alcalde, Jana Strnadov{\'a}, Esha Banerjee, Ruli Manurung, Antonio Stella, Atsuko Shimada,
  Sookyoung Kwak, Gustavo Mendon{\c{c}}a, Tatiana Lando, Rattima Nitisaroj, and Josie Li. 2017.
\newblock \href {https://doi.org/10.18653/v1/K17-3001} {{C}o{NLL} 2017 shared task: Multilingual parsing from raw text to {U}niversal {D}ependencies}.
\newblock In \emph{Proceedings of the {C}o{NLL} 2017 Shared Task: Multilingual Parsing from Raw Text to Universal Dependencies}, pages 1--19, Vancouver, Canada. Association for Computational Linguistics.

\bibitem[{Zhou et~al.(2023)Zhou, Zhang, Gu, Chen, and Poon}]{zhou2023universalner}
Wenxuan Zhou, Sheng Zhang, Yu~Gu, Muhao Chen, and Hoifung Poon. 2023.
\newblock \href {https://arxiv.org/pdf/2308.03279} {{UniversalNER: Targeted Distillation from Large Language Models for Open Named Entity Recognition}}.
\newblock \emph{arXiv preprint arXiv:2308.03279}.

\end{thebibliography}

\newpage

\appendix
\section{Contributions}

\noindent\textbf{Stephen Mayhew} conception, kickoff, all initial organization, recruitment, and annotation, development of annotation tool, manuscript writing.\vspace{2mm}\\
\textbf{Terra Blevins} annotation, organization, PUD analysis scripts, core manuscript writing.\vspace{2mm}\\
\textbf{Shuheng Liu} annotation, all baseline experiments and analysis.\vspace{2mm}\\
\textbf{Marek Šuppa} annotation, PROPN analysis, paper writing, GPU resources.\vspace{2mm}\\
\textbf{Hila Gonen} advising, organization of and feedback on manuscript.\vspace{2mm}\\
\textbf{Joseph Marvin Imperial} facilitated annotations for Tagalog and Cebuano, additions to manuscript for TL/CEB results, limitations, ethics, and conclusion sections.\vspace{2mm}\\
\textbf{Börje F. Karlsson} annotation, manuscript writing and editing, advising.\vspace{2mm}\\
\textbf{Peiqin Lin} annotation, manuscript comments.\vspace{2mm}\\
\textbf{Nikola Ljubešić} preparation and transfer of the HR \textsc{set} and SR \textsc{set} datasets, manuscript comments and edits.\vspace{2mm}\\
\textbf{LJ Miranda} annotation, related work section, comments, edits.\vspace{2mm}\\
\textbf{Barbara Plank} preparation and transfer of the DA \textsc{ddt} dataset, manuscript writing, comments and edits.\vspace{2mm}\\
\textbf{Arij Riabi} preparation and transfer of North African Arabizi dataset, related work section, comments and edits.\vspace{2mm}\\
\textbf{Yuval Pinter} advising, organization of and writing of manuscript.

\section{Additional Dataset Details}
In this section, we provide additional statistics and analysis of the datasets included in \shortname{} v1. Table \ref{tab:uner-domains} documents the domains included in each dataset along with their distributions of NER tags, and
Table \ref{tab:propn_overlap_f1} presents the F1 overlap score between named entities in \shortname{} and \textsc{propn} tags in the underlying UD treebanks. We also report the full numerical results of our baseline experiments in Table \ref{tab:baseline_results}.

\begin{table}[h]
\centering
\begin{tabular}{llrrr}
\toprule
Lang. & Dataset & Train & Dev & Test \\
\midrule
\textsc{da} & \texttt{ddt} & .824 & .835 & .836 \\
\textsc{en} & \texttt{ewt} & .813 & .815 & .817 \\
\textsc{hr} & \texttt{set} & .810 & .785 & .799 \\
\textsc{pt} & \texttt{bosque} & .844 & .859 & .856 \\
 \textsc{qaf} & \texttt{arabizi} & .942 & .962 & .971 \\
\textsc{sk} & \texttt{snk} & .848 & .783 & .771 \\
\textsc{sr} & \texttt{set} & .811 & .769 & .807 \\
\textsc{sv} & \texttt{talbanken} & .830 & .805 & .839 \\
\textsc{zh} & \texttt{gsd} & .700 & .696 & .720 \\
\textsc{zh} & \texttt{gsdsimp} & .695 & .695 & .719 \\

\midrule

\textsc{de} & \texttt{pud} & -- & -- & .785 \\
\textsc{en} & \texttt{pud} & -- & -- & .826 \\
\textsc{pt} & \texttt{pud} & -- & -- & .805 \\
\textsc{ru} & \texttt{pud} & -- & -- & .779 \\
\textsc{sv} & \texttt{pud} & -- & -- & .877 \\
\textsc{zh} & \texttt{pud} & -- & -- & .708 \\

\midrule

\textsc{ceb} & \texttt{gja} & -- & -- & .926 \\
\textsc{tl} & \texttt{trg} & -- & -- & .696 \\
\textsc{tl} & \texttt{ugnayan} & -- & -- & .723 \\

\bottomrule
\end{tabular}%
\caption{Comparing PROPN overlap (F1 scores).}
\label{tab:propn_overlap_f1}
\end{table}

\begin{table*}
    \centering
    \begin{adjustbox}{max width=\linewidth}
    \begin{tabular}{l ll l rrr}
    \toprule
    & & & & \multicolumn{3}{c}{Entity Dist. (\%)} \\
    \cmidrule(lr){5-7}
    Data Source & Lang. & Dataset & Domains & \texttt{LOC} & \texttt{ORG} & \texttt{PER}\\
    \midrule
    \citet{johannsen-etal-2015-universal} & \textsc{da} & \texttt{ddt} & fiction, news, nonfiction, spoken & 28.0 & 30.8 & 41.2 \\
    \citet{silveira-etal-2014-gold} & \textsc{en} & \texttt{ewt} & blog, email, reviews, social, web & 37.8 & 21.8 & 40.4 \\
    \citet{agic-ljubesic-2015-universal} & \textsc{hr} & \texttt{set} & news, web, wiki & 37.4 & 33.0 & 29.6 \\
    \citet{rademaker-etal-2017-universal} & \textsc{pt} & \texttt{bosque} & news & 29.5 & 33.9 & 36.6 \\
    \citet{seddah-etal-2020-building} & \textsc{qaf} & \texttt{arabizi} & blog, web, social & 57.5 & 27 & 15.4 \\
    \citet{zeman-2017-slovak} & \textsc{sk} & \texttt{snk} & fiction, news, nonfiction & 21.2 & 6.2 & 72.6 \\
    \citet{samardzic-etal-2017-universal} & \textsc{sr} & \texttt{set} & news & 41.4 & 30.2 & 28.4 \\
    \citet{mcdonald-etal-2013-universal} & \textsc{sv} & \texttt{talbanken} & news, nonfiction & 54.0 & 20.0 & 25.0 \\
    \citet{ud_chinese_gsd} & \textsc{zh} & \texttt{gsd} & wiki & 48.1 & 17.9 & 34.0 \\
    \citet{ud_chinese_gsdsimp} & & \texttt{gsdsimp} & wiki & 48.0 & 18.0 & 34.0 \\
     \midrule
     \multirow{6}{*}{\citet{zeman-etal-2017-conll}}& \textsc{de} & \texttt{pud} & news, wiki & 41.3 & 18.5 & 40.2 \\
     & \textsc{en} & \texttt{pud} & news, wiki & 39.5 & 21.9 & 38.6 \\
     & \textsc{pt} & \texttt{pud} & news, wiki & 44.4 & 18.0 & 37.6 \\
     & \textsc{ru} & \texttt{pud} & news, wiki & 33.4 & 26.6 & 40.0 \\
     & \textsc{sv} & \texttt{pud} & news, wiki & 43.0 & 15.7 & 41.3 \\
     & \textsc{zh} & \texttt{pud} & news, wiki & 44.9 & 13.5 & 41.6 \\
     \midrule    
     \citet{aranes-2022-gja} & \textsc{ceb} & \texttt{gja} & grammar examples & 12.3 & 2.0 & 85.7 \\
     \citet{ud_tagalog_trg} & \textsc{tl} & \texttt{trg} & grammar examples & 10.9 & 0.0 & 89.1 \\
     \citet{ud_tagalog_ugnayan} & & \texttt{ugnayan} & fiction, nonfiction & 47.5 & 0.0 & 52.5 \\
    \bottomrule
    \end{tabular}
    \end{adjustbox}
    \caption{Domains and distribution of entity types for datasets in UNER. Domains are categorized for the underlying UD datasets at \url{https://universaldependencies.org/}.}
    \label{tab:uner-domains}
\end{table*}

\begin{table*}[ht]
    \centering
    \begin{tabular}{llllrrrr}
         \toprule
         \multicolumn{2}{c}{Source} & \multicolumn{2}{c}{Target } & \multicolumn{4}{c}{F\textsubscript{1}}\\
         \cmidrule(lr){1-2} \cmidrule(lr){3-4} \cmidrule(lr){5-8}
         Lang. & Dataset & Lang. & Dataset & \texttt{LOC} &\texttt{ORG} & \texttt{PER} & Overall \\
         \midrule
         \textsc{da} & \texttt{ddt} & \textsc{da} & \texttt{ddt} & .879 & .826 & .924 & .878 \\
        \textsc{en} & \texttt{ewt} & \textsc{en} & \texttt{ewt} & .871 & .709 & .950 & .858 \\
        \textsc{hr} & \texttt{set} & \textsc{hr} & \texttt{set} & .977 & .891 & .970 & .950 \\
        \textsc{sk} & \texttt{snk} & \textsc{sk} & \texttt{snk} & .846 & .635 & .882 & .855 \\
        \textsc{pt} & \texttt{bosque} & \textsc{pt} & \texttt{bosque} & .882 & .861 & .966 & .904 \\
        \textsc{qaf} & \texttt{arabizi} & \textsc{qaf} & \texttt{arabizi} & .780 & .520 & .717 & .711 \\
        \textsc{sr} & \texttt{set} & \textsc{sr} & \texttt{set} & .981 & .913 & .983 & .962 \\
        \textsc{sv} & \texttt{talbanken} & \textsc{sv} & \texttt{talbanken} & .904 & .742 & .928 & .883 \\
        \textsc{zh} & \texttt{gsd} & \textsc{zh} & \texttt{gsd} & .906 & .819 & .922 & .896 \\
        \textsc{zh} & \texttt{gsdsimp} & \textsc{zh} & \texttt{gsdsimp} & .906 & .802 & .925 & .894 \\
        \midrule
        \multirow{6}{*}{\textsc{en}} & \multirow{6}{*}{\texttt{ewt}} & \textsc{de} & \texttt{pud} & .816 & .603 & .893 & .814 \\
        && \textsc{en} & \texttt{pud} & .785 & .593 & .922 & .805 \\
        & & \textsc{pt} & \texttt{pud} & .845 & .698 & .914 & .848 \\
        & & \textsc{ru} & \texttt{pud} & .681 & .451 & .875 & .715 \\
        & & \textsc{sv} & \texttt{pud} & .887 & .655 & .928 & .872 \\
        & & \textsc{zh} & \texttt{pud} & .465 & .308 & .389 & .410 \\
        \midrule
        \multirow{3}{*}{\textsc{en}} & \multirow{3}{*}{\texttt{ewt}} & \textsc{ceb} & \texttt{gja} & .556 & .000 & .842 & .789 \\
        && \textsc{tl} & \texttt{trg} & 1.00 & -- & .923 & .936 \\
        && \textsc{tl} & \texttt{ugnayan} & .857 & -- & .000 & .783 \\
                 \bottomrule
    \end{tabular}
    \caption{The full results of our baseline experiments from finetuning XLM-R\textsubscript{Large} on UNER. All scores are reported in micro-F\textsubscript{1}. \texttt{ORG} F\textsubscript{1} scores are not reported for the two \textsc{TL} datasets since there are no \texttt{ORG} entities labeled.}
    \label{tab:baseline_results}
\end{table*}

\section{Dataset Licensing}

The Universal Dependencies datasets are licensed under Creative Commons Attribution-ShareAlike (CC BY-SA)\footnote{\url{creativecommons.org/licenses/by-sa/4.0/}}. 
This license requires that ``if you remix, transform, or build upon the material, you must distribute your contributions under the same license as the original.'' Thus, we distribute all of our datasets under the same license.

\end{document}